\documentclass{WileyChemistry-template}
\usepackage{glossaries}
\usepackage{makecell}
\usepackage{subcaption}
\usepackage{chemformula}
\usepackage{amsfonts,amssymb}
\usepackage{dsfont}
\usepackage{threeparttable}
\usepackage{lineno}
\usepackage{color}
\usepackage{csquotes}
\crefname{subsection}{subsection}{subsections}
\Crefname{subsection}{Subsection}{Subsections}

\ifdefined\REVIEW
   \linenumbers
   
\fi

\title{\raggedright Accelerating battery research with an interoperable interface between FINALES and Kadi4Mat}

\author{
\begin{minipage}{\textwidth}
	Giovanna Tosato,*\textsuperscript{[a]} Leon Merker,\textsuperscript{[c, d, e, f, g, h]} Monika Vogler,\textsuperscript{[c]} Michael Selzer,\textsuperscript{[l]} Arnd Koeppe,*\textsuperscript{[a,b]}
\end{minipage}
}

\newcommand{\affiliation}{
\begin{itemize}		    

\item[{[a]}]
Institute for Applied Materials - Microstructure Modeling and Simulation (IAM-MMS), 
Karlsruhe Institute of Technology (KIT), Germany

\item[{[b]}]
Institute of Nanotechnology - Microstructure Simulation (INT-MSS)
Karlsruhe Institute of Technology (KIT), Germany

\item[{[c]}] 
Helmholtz Institute Ulm (HIU), Karlsruhe Institute of Technology (KIT), Germany

\item[{[d]}] 
TUM School of Natural Sciences, Department of Chemistry, Chair of Digital Catalysis, Munich

\item[{[e]}]
Institute of Robotics and Machine Intelligence (MIRMI), Technical University of Munich, Germany

\item[{[f]}]
Munich Data Science Institute (MDSI), Germany

\item[{[g]}]
Munich Institute of Integrated Materials, Energy and Process Engineering (MEP), Germany

\item[{[h]}]
Munich Center for Machine Learning (MCML), Germany

\item[{[i]}]
Technical University of Munich, TUM Department of Chemistry, Germany

\item[{[l]}]
Institute of Nanotechnology - Research Data Management (INT-RDM), Karlsruhe Institute of Technology (KIT), Germany

\end{itemize}
}

\renewcommand{\abstract}{
This study investigates how automated and interoperable research infrastructures can accelerate experimental materials research, using battery formation as a case study. 
We introduce a methodological framework that integrates the \gls{FINALES} and Kadi \gls{RDM} ecosystems, enabling coordinated experiment execution, data management, and analysis across distributed research infrastructures.

The \gls{FINALES} framework orchestrates experiment planning and execution on the \gls{POLiS} \gls{MAP}, while Kadi handles data management, visualization, experiment selection, and persistent storage.
This interoperability enables reproducible, coordinated, end-to-end experimental workflows that connect automated systems and human-operated processes across multiple research sites.

To showcase the framework's capabilities, we investigate the formation process of sodium-ion coin cells, a critical step that influences cell lifetime.
Using a dataset generated through the proposed infrastructure, we analyze the relationships between formation parameters and electrochemical performance.
A Gaussian Process model is employed to identify regions associated with favorable performance within the investigated parameter domain, while quantifying predictive uncertainty, highlighting where additional experimental evidence would be most informative.

The presented workflow demonstrates that interoperable, automated infrastructures can support experimental campaigns, enhance data consistency and traceability, and support reproducible data-driven analysis.
While demonstrated through a battery application, the framework provides a general methodology for integrating distributed experimental platforms with \gls{RDM} systems, transferable across materials science and engineering domains.
}

\newcommand{\keywords}{
	End Of Life \textbullet\ 
	FINALES \textbullet\ 
	Kadi4Mat \textbullet\ 
	Materials Acceleration Platform
}

\newacronym{MAP}{MAP}{Materials Acceleration Platform}
\newacronym{FINALES}{FINALES}{Fast-INtention Agnostic LEarning Server}
\newacronym{OVERLORT}{OVERLORT}{OVERLooking ORchestrating Tenant}
\newacronym{SDL}{SDL}{Self-Driving Lab}
\newacronym{EOL}{EOL}{End Of Life}
\newacronym{CIDS}{CIDS}{Computational Intelligence and Data Science framework}
\newacronym{SEI}{SEI}{Solid-Electrolyte Interface}
\newacronym{SI}{SI}{Supporting Information}
\newacronym{POLiS}{POLiS}{Post-Lithium Storage Cluster of Excellence}
\newacronym{RDM}{RDM}{Research Data Management}
\newacronym{GP}{GP}{Gaussian Process}
\newacronym{SEM}{SEM}{Standard Error of the Mean}
\newacronym{UUID}{UUID}{Universally Unique Identifier}
\newacronym{RBF}{RBF}{Radial Basis Function}
\newacronym{ARD}{ARD}{Automatic Relevance Determination}
\newacronym{RQ}{RQ}{Qational Quadratic}
\newacronym{AutoBASS}{AutoBASS}{Automatic Battery ASsembly System}

\DeclareSIUnit{\molr}{\textsc{m}}

\begin{document}

\twocolumn[\vspace{-1.5cm}\maketitle\vspace{-1cm}
	\textit{\dedication}\vspace{0.4cm}]
\small{\begin{shaded}
		\noindent\abstract
	\end{shaded}
}

\begin{figure} [!b]
\begin{minipage}[t]{\columnwidth}{\rule{\columnwidth}{1pt}\footnotesize{\textsf{\affiliation}}}\end{minipage}
\end{figure}



\section{Introduction}
\label{introduction}
Rapid advances in battery technology are constrained by the time- and resource-intensive nature of cell conditioning. 
The battery production chain involves several stages, including material preparation, electrode production, cell manufacturing, and cell conditioning \cite{kwade2018current}. 
Among these, cell conditioning remains one of the most time- and resource-intensive phases, which significantly slows down innovation. 
Notably, the longest steps that occur during the conditioning phase are the formation and aging processes, which can last for up to 24 hours \cite{heimes2018lithium} and three weeks \cite{heimes2020effects, wood2019formation}, respectively. 
Formation establishes the electrochemical interfaces that largely determine long-term cell performance. During aging, quality and safety parameters are monitored.
The associated costs of the formation and aging processes account for approximately \SI{14}{\percent} \cite{nelson2019modeling, borner2024prozessbasiertes, hakimian2015economic} to \SI{30}{\percent} \cite{kwade2018current, liu2021current} of the total cell production expenses. 
Part of these costs is the necessary equipment, which requires a significant amount of floor space, up to \SI{25}{\percent} of the entire production facility \cite{wood2019formation}. 
Formation procedures are highly material-dependent and rarely directly transferable to alternative materials. 
Furthermore, scaling observations from small lab cells (e.g., coin cells) to large industrial formats is challenging because new physical behaviors emerge \cite{scurtu2025small}.
For these reasons, formation remains one of the least standardized and most critical bottlenecks in battery manufacturing, particularly for emerging chemistries such as sodium-ion batteries, where comparatively little experimental data are available \cite{CUI20243072, STOCK2024234858}.
Approaches in \glspl{MAP} have demonstrated the potential of laboratory automation to accelerate discovery and increase experimental throughput\cite{vogler2023brokering, vogler2024autonomous, dave2020autonomous,wagner2021evolution,walter2010combinatorial,burger2020mobile}. 
However, \glspl{MAP} that focus on investigating material properties or reaction pathways, rarely account for device-level processes such as formation, which remain difficult to integrate.
The \gls{FINALES} framework combines both and was demonstrated in an investigation of electrolyte formulations regarding their ionic conductivity and \gls{EOL}\cite{vogler2024autonomous, vogler2023brokering}. 
The \gls{EOL} is an indicator for the maximum application duration commonly described by the number of cycles, after which the capacity of the cell falls below \SIrange{70}{80}{\percent} of its initial capacity \cite{wang2022prospects, chen2019recycling, ramoni2013end, saxena2015quantifying}. 
A further aspect to consider is that fully autonomous workflows are rarely achievable in practice at device level \cite{Hung2024Autonomous, Gary2024SelfDriving}, as important experimental tasks continue to require human intervention due to technical limitations, safety \cite{scheurer2025role}, or practical constraints.
Consequently, future experimental frameworks must support seamless interaction between automated equipment, \gls{RDM} systems, and human-operated routines.
Equally important is the ability of such frameworks to bridge collaborating research groups that contribute to different stages of the experimental data life cycle.

At the same time, the increasing availability of automated experimentation shifts the research bottleneck from data generation to data management and interpretation.
Experimental campaigns often involve heterogeneous software environments and independent research infrastructures, limiting data interoperability, traceability, and reuse.
Modern battery research is no exception, relying on heterogeneous \gls{RDM} systems to store, process, and exchange experimental data.
However, limited interoperability between such systems often hinders efficient reuse of data, coordination, and automation across platforms.
This work addresses these methodological challenges in distributed experimental environments by establishing interoperability between two complementary \gls{RDM} ecosystems: Kadi and the \gls{FINALES}-based \gls{POLiS} \gls{MAP}.
The Kadi ecosystem consists of multiple instances and modules, among which Kadi4Mat serves as the virtual research environment that manages data, metadata, and workflow automation. 
It combines repository functionality for data management with electronic lab notebook capabilities and a plugin system, enabling the documentation and automation of research processes.
The FINALES framework orchestrates the planning and execution of experiments on the \gls{POLiS} \gls{MAP} \textit{via} so-called tenants, which combine automated capabilities with tasks performed by human researchers.
We bridge these systems by establishing interoperable links \textit{via} dedicated plugins and tenants (see \cref{fig_concept}).

The capabilities of the resulting interactive framework are demonstrated through the investigation of formation protocols for sodium-ion coin cells.
We generate a cell dataset by deploying automated laboratory experimentation coupled with human-in-the-loop process steps through the presented framework. 
Furthermore, we employ \gls{GP} modeling to characterize the relationship between formation conditions and electrochemical performance, identify promising operating conditions, and quantify predictive uncertainty throughout the investigated parameter domain.

Accelerating battery development has become increasingly important as demand for cost-efficient, sustainable energy storage technologies continues to grow.
The presented framework illustrates how interoperable automation and \gls{RDM} can accelerate experimental battery research while providing a transferable methodology for distributed materials research beyond the application considered in this work.


\section{Results and Discussion}
\label{sec:results}
The primary contribution of this work is the development of an interoperable framework that coordinates experiment planning, execution, analysis, and FAIR data management across distributed research infrastructures.
The framework presented in \cref{subsec:method-results} demonstrates interoperability between two complementary \gls{RDM} systems through the Kadi-FINALES coupling.
This interoperability can improve operational efficiency by integrating automated laboratories and human-operated workflows, enhancing data exchange, reproducibility, and reuse.
In addition, \cref{subsec:application-results} illustrates the capabilities of the proposed methodology through an experimental campaign on sodium-ion coin cells.
The resulting dataset, covering a range of formation conditions, is subsequently analyzed using \gls{GP} modeling to investigate the relationship between formation parameters and electrochemical performance. 
This analysis aims to identify promising operating conditions within the investigated domain while quantifying predictive uncertainty. 
Together, these results illustrate how interoperable research infrastructures can accelerate experimental studies while providing a robust foundation for data-driven analysis.

\subsection{Methodological contribution: a human-in-the-loop interoperable framework}
\label{subsec:method-results}
In this subsection, we follow the data flow between the connected processes involved in the study, from the operating user to the \gls{MAP} and back.
By combining the complementary functionalities of Kadi and \gls{FINALES}, the human-in-the-loop experimental framework supports the design, execution, and analysis of battery experiments by integrating \gls{RDM}, automated laboratory infrastructure, and human expertise.

\cref{fig_concept} shows the workflow sequence at a conceptual level. 
The workflow begins in Kadi, where the user defines the study scope through Kadi4Mat using a user-friendly template.
This records the parameters of interest, their admissible ranges, and any relevant constraints.
Based on this configuration, candidate experimental trials within the specified domain can be requested. 
The framework supports multiple strategies for selecting such candidates, ranging from manually predefined to pseudo-random and data-driven optimization schemes.
Specifically, when the objective is to optimize one or more performance metrics, an active learning strategy based on Bayesian optimization can be employed for candidate selection.
The selection strategy remains a configurable component of the framework, while the enclosing workflow for experiment execution and data management remains independent from the choice.
The resulting experimental configurations are then submitted to the \gls{FINALES} server, which translates it into executable workflow requests, thereby configuring the corresponding experiments for lab execution.
Once all results are available, they are automatically uploaded to Kadi4Mat in a predefined data structure and incorporated into the study dataset.
Technical details of the Kadi components are provided in \cref{subsec:methods-kadi}, and details of the FINALES components are provided in \cref{subsec:methods-finales}.

\begin{figure*}[bthp]
\begin{center}
\includegraphics[width=17.4cm]{new_figs/concept.jpg}
\caption{The conceptual framework that interfaces the Kadi ecosystem with the \gls{FINALES} workflow system for automated experiments on the \gls{POLiS} \gls{MAP}. 
In the Kadi ecosystem, the user defines experiment specifications and triggers the workflow through Kadi4Mat. The \gls{FINALES} plugin converts the passed parameter configurations into a workflow request. Through \gls{FINALES}, the \gls{OVERLORT} orchestrates the individual services to execute the workflow request in the automated laboratory and uploads results back to Kadi4Mat through the Kadi tenant.}
\label{fig_concept}
\end{center}
\end{figure*}

The web interface of Kadi4Mat, provides an intuitive way to access and interact with the data, enhancing the user-friendliness and navigability of the data generated within the \gls{POLiS} \gls{MAP}.
Moreover, the connection to Kadi4Mat not only provides data visualization tools but also enables researchers to add observations that automated procedures may miss.
In this demonstration, Kadi4Mat provided an interface for human researchers to add descriptions, comments, or observations, allowing proper handling of affected data points and increased traceability for decisions such as outlier detection.
Furthermore, maintaining accessibility for human researchers opens the possibility of integrating non-automated laboratories into \glspl{MAP} running on \gls{FINALES}.
This enables a more inclusive research strategy where automated and human-operated processes can complement one another, leveraging the versatility of human researchers where full automation is impractical or undesirable \cite{Gary2024SelfDriving}.
At the same time, standardized automated procedures and their associated code bases can be shared and reused, making efficient use of the invested development resources and improving the reproducibility of experiments.

Since Kadi4Mat serves a central role in communication and information exchange with the \gls{POLiS} \gls{MAP}, all data associated with requests and corresponding results are recorded there.
Data are shared with all project partners, enabling continuous access and evaluation throughout the study.
Furthermore, publishing data for a larger community is significantly easier than storing all data locally and then manually uploading them at the end of the study.
Kadi4Mat provides means to export the data and associated metadata of a record and directly publish them in external repositories such as Zenodo.

\subsection{Experimental Validation: formation time and life cycle of Sodium ion coin-cell}
\label{subsec:application-results}
To demonstrate the presented methodology end-to-end, we generated an experimental dataset of sodium-ion coin cells varying the formation protocol and analyzing the resulting cell behavior.
The formation protocol was parameterized by the charge and discharge rate applied during formation, and the number of formation cycle repetitions (\cref{tab:search_space}). 
The analyzed dataset comprises experimental configurations from two sources. 
Batches 0 to 10 originate from a seeding experimental study, in which human operators manually selected parameter configurations based on their prior knowledge. 
Batches 11 to 17 were generated using multi-objective Bayesian optimization, in which an evaluation script processed the raw cell data and returned the corresponding performance metrics as a black-box function.
Due to a software bug in the evaluation script, only a subset of the experimental data from each batch was initially evaluated. 
This caused incorrect values to be supplied to the optimizer, breaking systematic optimization and leading to a `pseudo-random' coverage of the design space.
Although the optimization algorithm and the overall framework executed correctly within the constraints of the flawed black-box function, experimental decisions were consequently informed by incorrect evaluations of the preceding experiments.
The evaluation-script bug has since been corrected. The individual cycling and performance data were fully recovered and used for further analysis. 
Both performance metrics improved over the course of the study; however, the incorrect evaluations likely affected the efficiency of the experimental campaign and may have limited the performances ultimately achieved.

For each cell, two descriptors were analyzed: the total formation time and the \gls{EOL} cycle number, defined as the number of cycles endured before the capacity falls below \SI{80}{\percent} of the initial discharge capacity measured in the first cycle after formation.
The dataset comprises a total of 18 batches, indexed from 0 to 17 (see \cref{tab_param_and_results}).
To improve resource utilization and operational efficiency, we evaluated up to three batches in parallel in each execution.
Furthermore, each batch comprised four cells assembled and cycled under the same protocol to reduce errors and minimize the impact of potential outliers.
The reported metrics correspond to the mean values obtained across these four cells, together with the \gls{SEM}, so that cell-to-cell variability is carried explicitly through the analysis.
One cell in batch 16 malfunctioned, and one cell in batch 9 failed immediately after the onset of regular cycling; both cells were treated as outliers and excluded from the calculations. 
Details on the experimental setups can be found in \cref{subsec:methods-experimental}.

\begin{table}
	\begin{center}
	\caption{Ranges defining the parameter space selected for the study.}
    \label{tab:search_space}
		\begin{tabular}{lcc}	
\toprule		
Parameter & range & \\
\midrule
C-rate charge formation & $[0.025, 2.005]$\\
C-rate discharge formation & $[0.025, 2.005]$\\
repetitions formation cycle & $[1, 6]$\\
\bottomrule
	\end{tabular}
	\end{center}
\end{table}

\begin{table*}[bthp]
	\begin{center}
	\caption{Formation protocols and measured descriptors for all the 18 batches generated during the study. The reported values of formation time and \gls{EOL} represent the mean of the four cells constituting each batch, with the corresponding \gls{SEM}. Batches marked in bold are non-dominated with respect to the pair
    (short formation time, high \gls{EOL}).}\label{tab_param_and_results}
	\begin{tabular}{lccc S[table-format=2.2] @{~${}\pm{}$~} S[table-format=2.2] S[table-format=3.2] @{~${}\pm{}$~} S[table-format=2.2]}
    Batch & \makecell{C-rate charge\\formation / \textrm{C}} 
    & \makecell{C-rate discharge\\formation / \textrm{C}} 
    & repetitions 
    & \multicolumn{2}{c}{\makecell{formation time\\$\pm$ \gls{SEM} / $\textrm{h}$}} 
    & \multicolumn{2}{c}{\makecell{EOL\\cycle number\\$\pm$ \gls{SEM}}}\\
    \midrule
    \textbf{0} & \textbf{1.50} & \textbf{1.50} & \textbf{3} & \textbf{1.74} & 0.03 & \textbf{359.75} & 98.42\\
    1 & 0.50 & 0.50 & 4 & 8.23 & 0.12 & 375.25 & 25.23\\
    2 & 1.00 & 1.00 & 6 & 4.28 & 0.28 & 303 & 123.53\\
    3 & 0.50 & 0.50 & 2 & 5.29 & 0.10 & 328.5 & 39.21\\
    4 & 1.00 & 1.00 & 3 & 2.89 & 0.03 & 421 & 85.91\\
    5 & 0.10 & 0.10 & 3 & 41.18 & 0.57 & 294 & 19.17\\
    6 & 0.18 & 0.18 & 2 & 15.74 & 0.09 & 120.25 & 41.59\\
    7 & 2.00 & 2.00 & 3 & 1.21 & 0.01 & 79.75 & 33.96\\
    8 & 0.16 & 0.16 & 4 & 32.41 & 0.39 & 76 & 17.33\\
    9 & 0.05 & 0.05 & 1 & 28.61 & 4.45 & 237.67 & 8.7\\
    10 & 0.10 & 0.10 & 1 & 15.81 & 0.11 & 181.75 & 51.06\\
    11 & 0.56 & 1.12 & 4 & 5.88 & 0.10 & 153.67 & 10.65\\
    12 & 1.15 & 0.60 & 4 & 3.92 & 0.67 & 182.25 & 45.20\\
    \textbf{13} & \textbf{1.33} & \textbf{1.25} & \textbf{1} & \textbf{0.90} & 0.17 & \textbf{221} & 99.37\\
    14 & 0.75 & 0.35 & 5 & 10.61 & 0.39 & 287.50 & 168.22\\
    15 & 1.59 & 1.66 & 5 & 2.37 & 0.08 & 862 & 272.00\\
    \textbf{16} & \textbf{1.12} & \textbf{1.33} & \textbf{4} & \textbf{2.24} & 0.72 & \textbf{1368.67} & 317.12\\
    17$^\dagger$ & 0.76 & 0.83 & 4 & 5.58 & 0.03 & 1156.25 & 130.26\\
    \bottomrule
	\end{tabular}
	\end{center}
    \begin{tablenotes}
        \item{ $\dagger$ \small{Batch 17 was discontinued after reaching the allocated resources budget. Consequently, the reported \gls{EOL} represents a lower bound of the actual \gls{EOL}.}}
    \end{tablenotes}
\end{table*}

The study concluded once the allocated experimental time and instrument occupancy were exhausted. For this reason, batch 17 was discontinued before two cells had reached the end-of-life threshold. The reported values should therefore be regarded as lower bounds of their actual \gls{EOL}.

\begin{figure*}[bthp]
    \centering
    \includegraphics[width=0.8\linewidth]{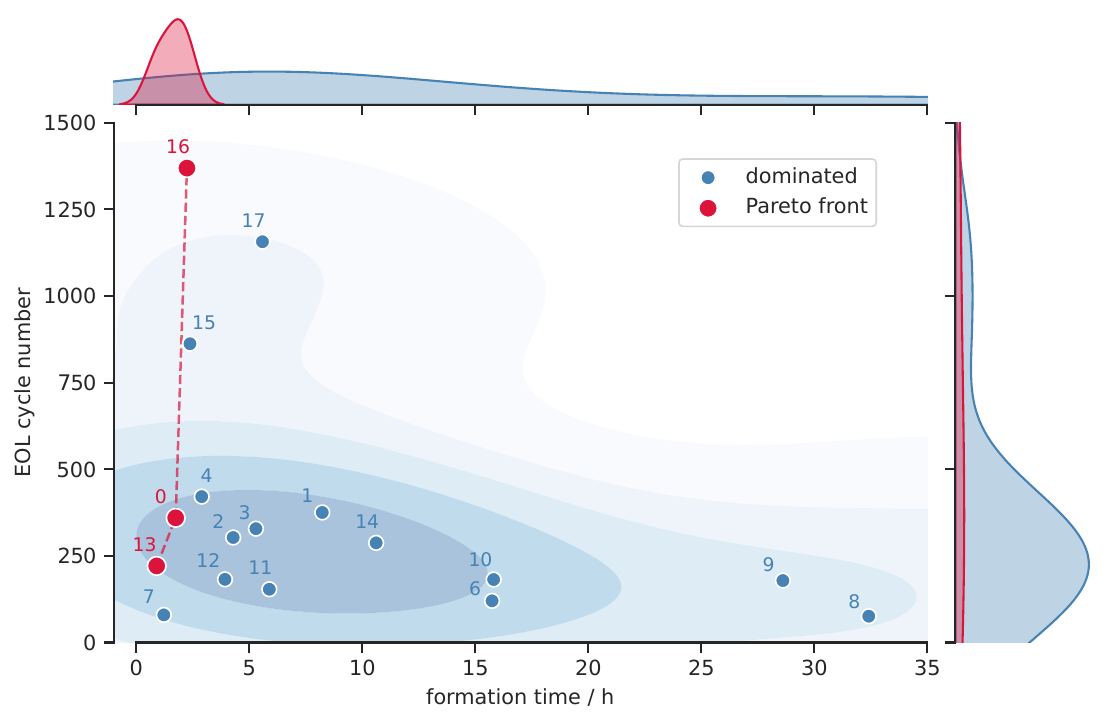}
    \caption{Visualization of the application results in the performance space. Batches 16, 0, and 13 lie on the Pareto front and represent good trade-off solutions identified in this study. These are highlighted in red, while the dominated solutions are in blue. 
    The marginal plots show the distributions of the observed values of formation time (on top), and \gls{EOL} (on the right), for both, Pareto points (in red) and dominated points (in blue). The observed data suggest that extended additional formation time does not buy extra durability.}
    \label{fig_PF}
\end{figure*}

Data and metadata associated with each cycling cell in these batches are stored in individual, mutually linked records in Kadi4Mat (cf. \cref{fig_data-knowledge-graph-kadi}), making the dataset directly reusable for modelling steps that follow.
\Cref{tab_param_and_results} provides an overview of the measurements obtained for each batch, along with the corresponding formation protocol. 
The resulting \gls{EOL} and formation time values for all samples are visualized in the performance space in \cref{fig_PF}.
Formation time spans from \SI{0.90}{\hour} (batch 13, a single formation cycle at \SI{1.33}{C}/\SI{1.25}{C}) to \SI{41.18}{\hour} (batch 5, three cycles at \SI{0.10}{C}) and is essentially dictated by the rates and the number of repetitions.
The \gls{EOL} cycle number varies from \num{76} (batch 8) to \num{1369} cycles (batch 16) and is not a monotonic function of either rate.
Within the scope of this experimental campaign, the longest-lived cells are concentrated in an intermediate band of rates: batches 16, 17 and 15, with rates between \SI{0.76}{C} and \SI{1.66}{C} and four to five formation repetitions, reach \num{1368.67}, \num{1156.25} and \num{862} cycles respectively, with formation times of \SI{2.24}{\hour}, \SI{5.58}{\hour} and \SI{2.37}{\hour}. 
This is consistent with the behavior reported for lithium-ion cells, where neither very long nor very short formation yields good long-term performance \cite{mao2018balancing}, and the present data indicate that a similar qualitative picture holds for sodium-ion coin cells.

\begin{figure*}[bthp]
    \centering
    \begin{subfigure}[b]{0.8\linewidth}
    \includegraphics[width=\linewidth]{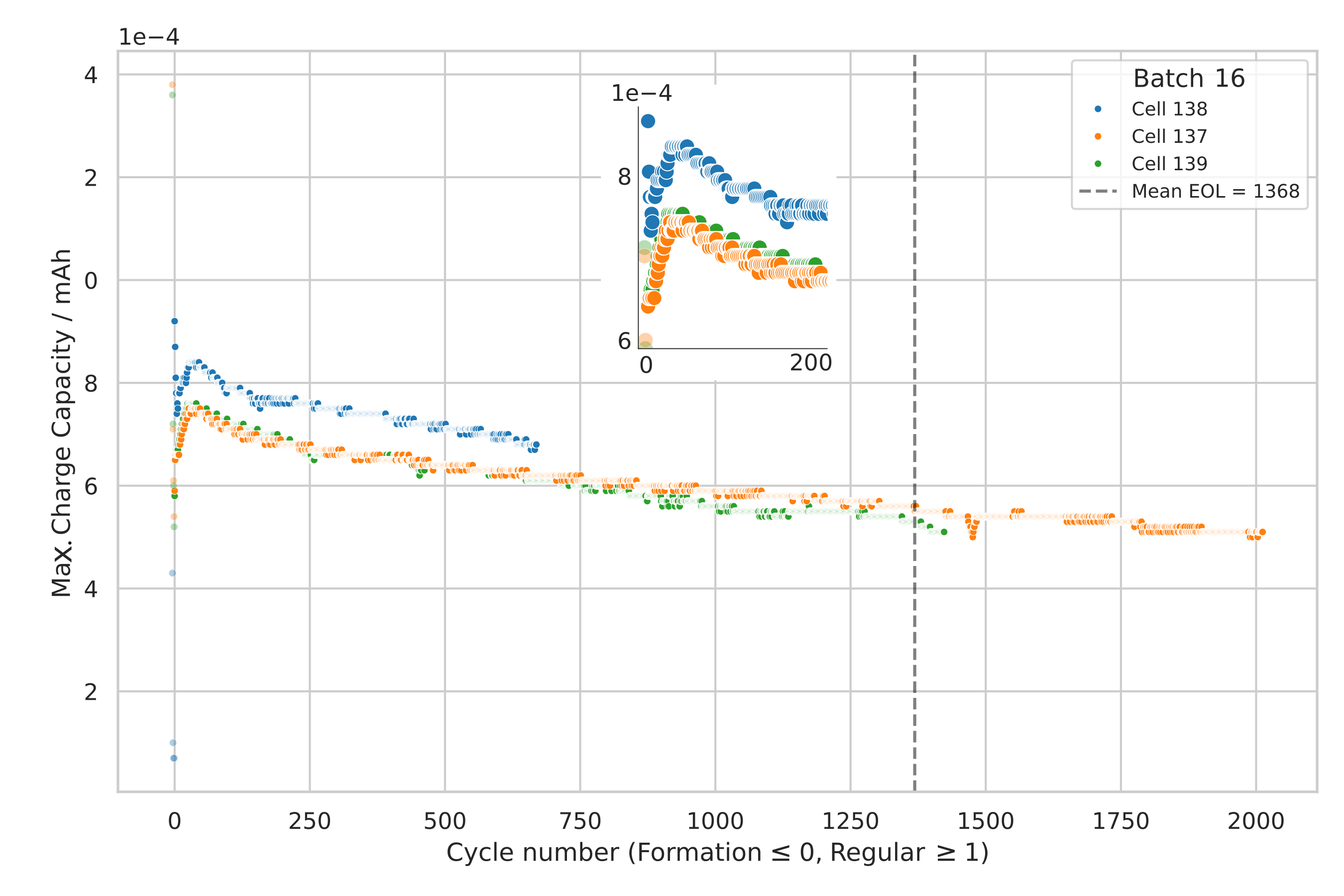}
    \caption{}\label{fig_charge_capacity_batch16}
    \end{subfigure}
    \begin{subfigure}[b]{0.8\linewidth}
    \includegraphics[width=\linewidth]{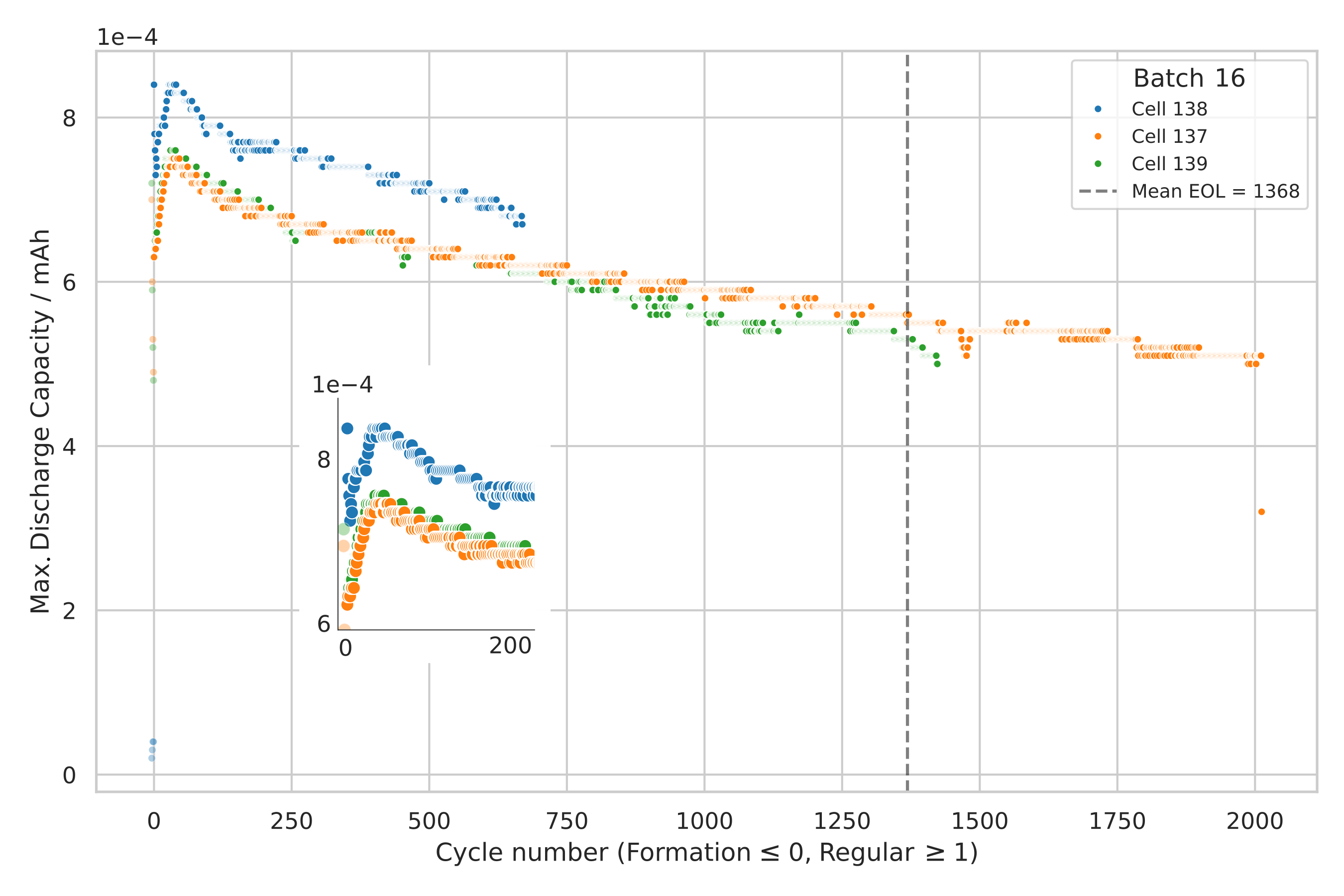}
    \caption{}\label{fig_discharge_capacity_batch16}
    \end{subfigure}
    \caption{Maximum charge (a) and discharge (b) capacity over cycles of the four cells constituting batch 16, which is a Pareto optimal solution within our findings. Batch 16 had one malfunctioning cell. Therefore, only 3 were considered for the analysis.}\label{fig_max_charge_discharge_capacity_batch16}
\end{figure*}

Considered jointly, three batches are non-dominated with respect to the pair (short formation time, high \gls{EOL}): batch 13 (\SI{0.90}{\hour}, \num{221} cycles), batch 0 (\SI{1.74}{\hour}, \num{359.75} cycles) and batch 16 (\SI{2.24}{\hour}, \num{1368.67} cycles). 
Batch 16 is the most informative of the three, combining the highest measured cycle life in the study with a formation stage of just over two hours.
Its \gls{SEM} (\num{317} cycles) is nevertheless large, as in other long-lived batches (\num{272} cycles for batch 15), indicating substantial cell-to-cell variation in the high-\gls{EOL} region. 
Maximum charge and discharge capacity over cycles for the cells of batch 16 are shown in \cref{fig_max_charge_discharge_capacity_batch16}.

Shifting from single-point observations to a description of the entire considered parameter domain, and aiming at an optimal compromise between low formation time and high \gls{EOL} performance, we employed \gls{GP} surrogates.
A \gls{GP} surrogate was fitted independently to each descriptor, since \gls{EOL} and formation time are physically distinct responses and, as shown below, they call for different covariance structures.
Each \gls{GP} surrogate is fitted to batch-level mean observations rather than to individual cell measurements. 
The batch mean is supplied as the observation, and the corresponding \gls{SEM} is passed to the model as known observation noise.
As a consequence, the predictive standard deviation reflects the model's uncertainty about the underlying trend, after accounting for the noise in the data.

\begin{figure*}[bthp]
    \centering
    \begin{subfigure}[b]{0.8\linewidth}
        \centering
        \includegraphics[width=0.7\textwidth]{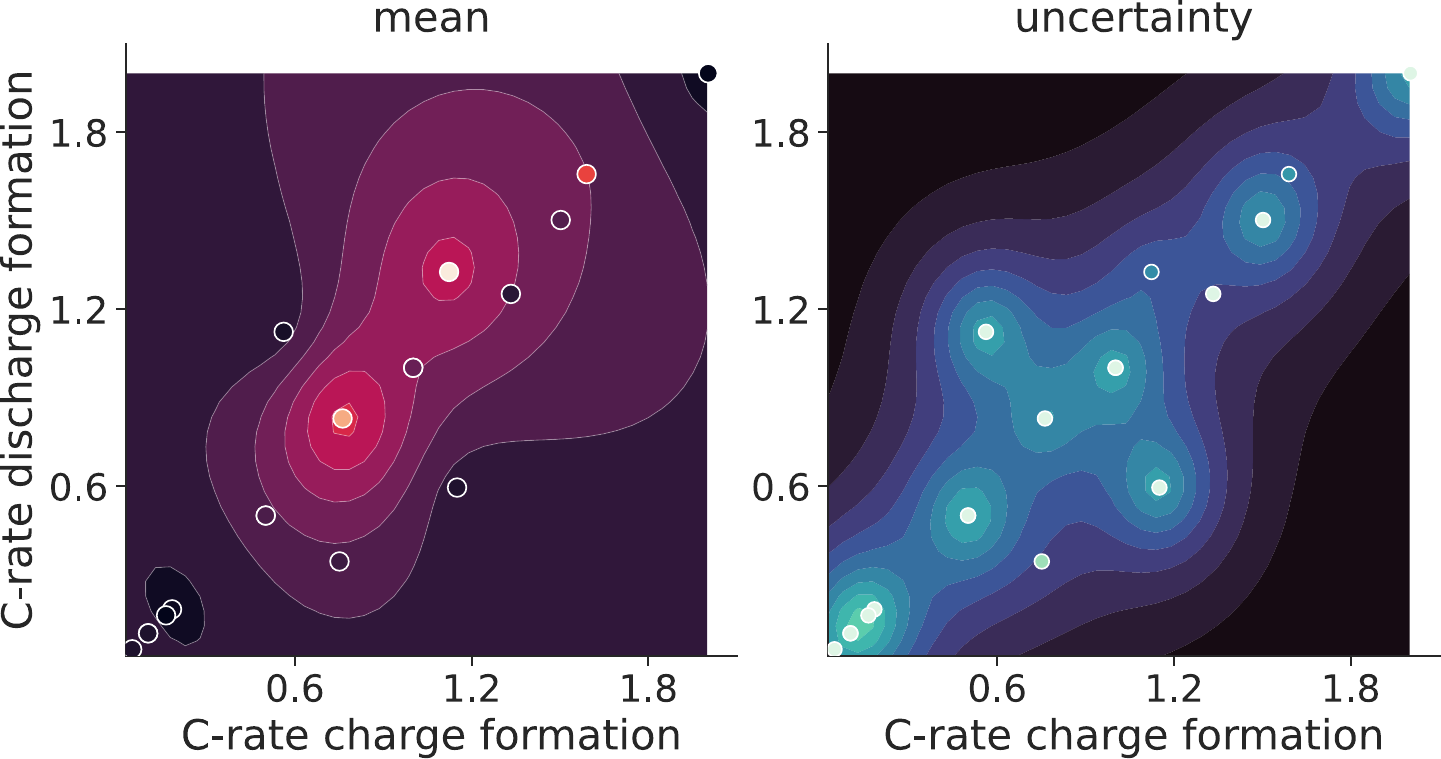}
        \caption{}\label{fig_EOL_contour_charge_discharge}
    \end{subfigure}
    \par\bigskip 
    \begin{subfigure}[b]{0.8\linewidth}
        \centering
        \includegraphics[width=0.7\textwidth]{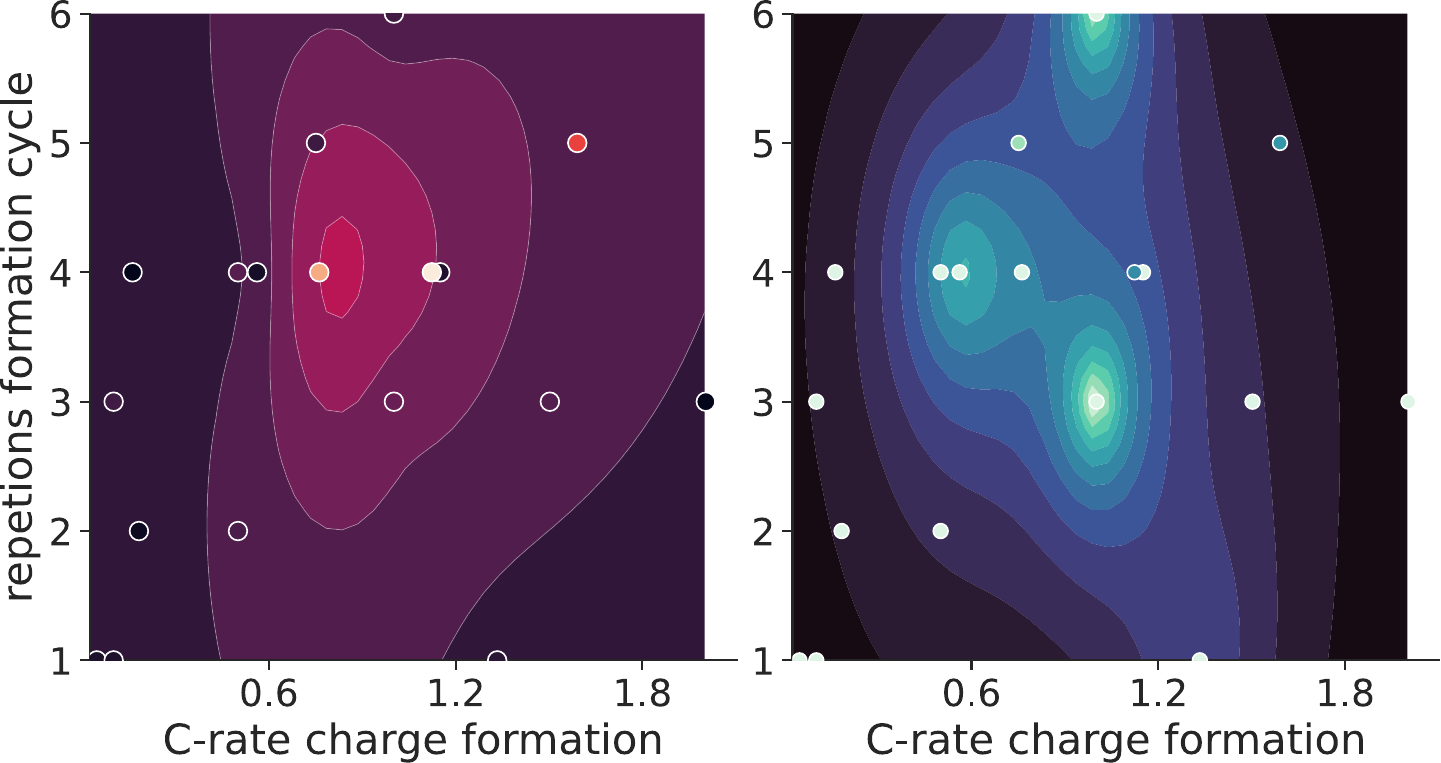}
        \caption{}\label{fig_EOL_contour_charge_rep}
    \end{subfigure}
    \par\bigskip 
    \begin{subfigure}[b]{0.8\linewidth}
        \centering
        \includegraphics[width=0.7\textwidth]{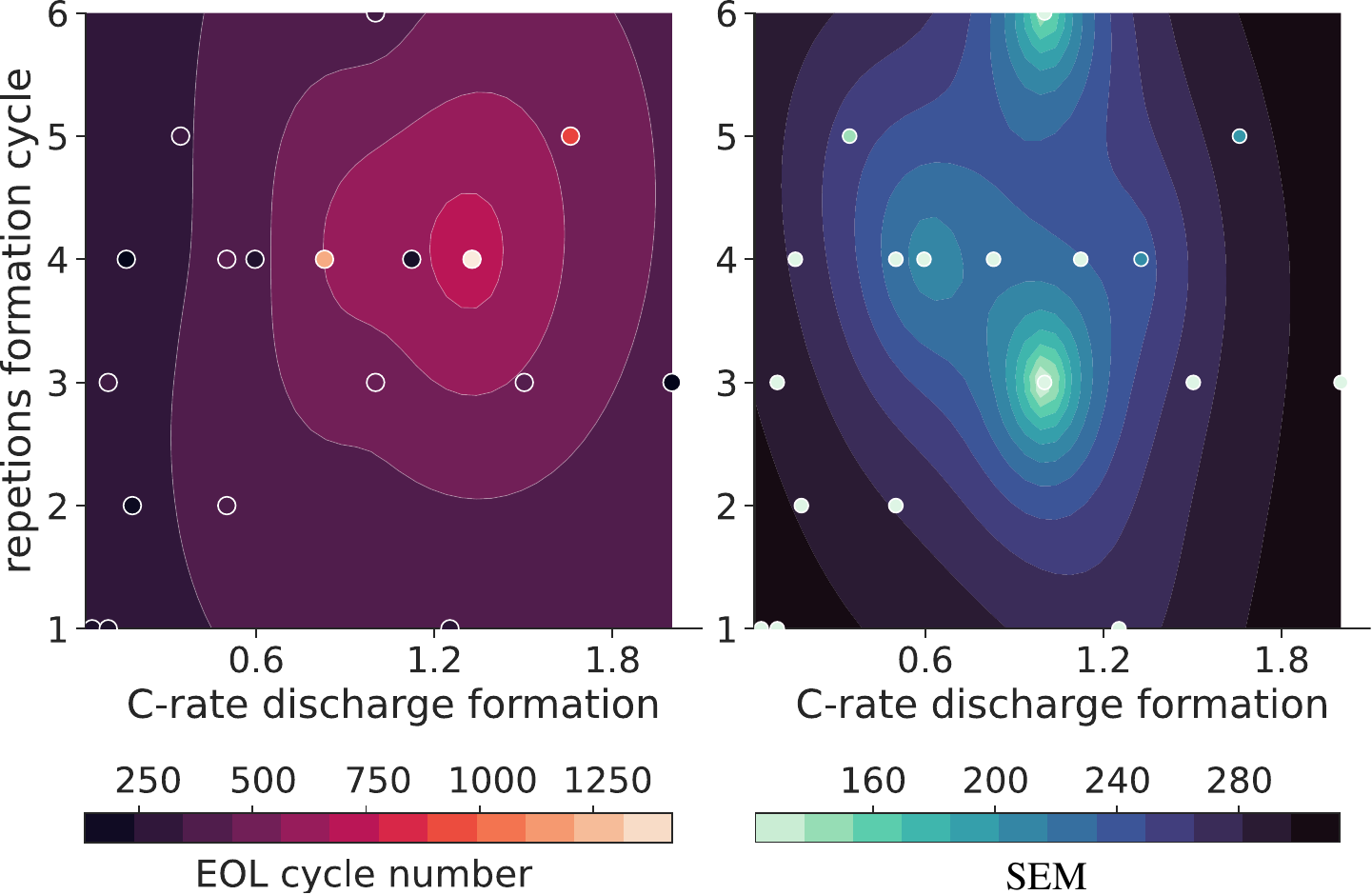}
        \caption{}\label{fig_EOL_contour_discharge_rep}
    \end{subfigure}
    \caption{Posterior mean and \gls{SEM} of the Matérn-$1/2$ \gls{GP} for \gls{EOL} cycle number across the investigated cycling-protocol parameter space. Contour plots show the modeled \gls{EOL} response for the three pairs of input dimensions: (a) charge and discharge rate, (b) charge rate and number of repetitions, (c) discharge rate and number of repetitions.} \label{fig_EOL_contours}
\end{figure*}

A \gls{GP} is uniquely described by its mean function and its covariance function (kernel).
Rather than committing to a single covariance function \textit{a priori}, we performed a kernel diagnostic study.
Six candidate kernels were considered, chosen to span a spectrum of assumed smoothness: the squared-exponential \gls{RBF} kernel, Matérn kernels with $\nu=5/2$, $\nu=3/2$ and $\nu=1/2$, the \gls{RQ} kernel, and an isotropic Matérn-$5/2$ kernel.
The smoothness of a kernel determines how rapidly the modeled function can vary: smoother kernels imply that nearby inputs tend to produce similar outputs, resulting in gradual changes in the predictions, whereas less smooth kernels allow for more abrupt variations between neighboring points.
Within the covariance function, the parameter controlling how fast the function changes over distance is called a lengthscale.
All kernels except the isotropic one use \gls{ARD}, allowing for one lengthscale per input dimension instead of a single lengthscale shared across all three. 
This way the fitted lengthscales are themselves a diagnostic: a large value indicates that the response varies slowly along that dimension, and hence that the descriptor is comparatively insensitive to it.
The formation protocol investigated in this study is jointly defined by the three considered input parameters: charge C-rate, discharge C-rate, and number of formation-cycle repetitions (for the details refer to \cref{subsec:methods-experimental}).
The kernel diagnosis returned that cycle life responds roughly to the formation protocol, meaning that neighboring protocols can differ sharply in the resulting descriptor values. 
In contrast, formation time is a smooth, near-deterministic function of the protocol. 
Therefore, naively fitting the same \gls{GP} for both objectives would have caused missing real local variations of \gls{EOL}, or introduced fake oscillations in an almost deterministic relationship for formation time. 
Based on the available data, we fit a Matérn-$1/2$ (exponential) kernel to describe \gls{EOL}, and an RBF kernel to describe formation time, i.e.
\begin{align*}
    f_{\text{EOL}}(x) &\sim GP(\mu_{\text{EOL}}, k_{\text{Matérn}1/2}) \\
    f_{\text{formation}}(x) &\sim GP(\mu_{\text{formation}}, k_{\text{RBF}}).
\end{align*}
This distinction is reflected in the fitted lengthscales, which indicate a similar sensitivity of the two objectives to the C-rates, but a marked difference in the sensitivity to the number of formation-cycle repetitions.
This suggests that this parameter is the primary driver of the difference in smoothness between \gls{EOL} and formation time.
\cref{fig_EOL_contours} shows the posterior mean and \gls{SEM} of the \gls{GP} with the selected kernel, for the \gls{EOL} cycle number projected onto the three pairs of input dimensions.

The model places the region of highest expected cycle life at intermediate to moderately high rates, for both charge and discharge, combined with around four formation repetitions.
For formation time, the posterior is monotonically decreasing in both rates (cf. \cref{fig_formation_contour} in \gls{SI}).
The combination of the two surfaces suggests that the short-formation region and the long-life region overlap only in a band around \SIrange{1.1}{1.6}{C}, which is where the non-dominated batches 0, 13 and 16 sit.

The predictive standard deviation is equally informative. For formation time it is small and roughly uniform, so little further evidence is required to characterize that response (cf. \cref{tab_param_and_results} and \cref{fig_formation_contour}).
For the \gls{EOL} cycle number, the SE is large across most of the domain and largest precisely in the region where the posterior mean suggests favorable performance. This is because only a few batches support this area, and their cell-to-cell variability is substantial.
This is where additional experimental evidence would be most valuable, making it a natural target for a follow-up campaign.
\cref{fig_boxplots} provides a more detailed view of cell-to-cell variability within each batch for both objectives and visually highlights batches with narrow or dispersed distributions.

By analyzing the dataset produced by the automated workflow, we captured the qualitative structure of the investigated formation-protocol domain, located the region associated with favorable cell performance, and explicitly quantified the model's uncertainty.
These insights into the electrochemical behavior of sodium-ion coin cells were made possible by the full traceability of the experimental dataset, which remained usable despite the software bug reported above.
This experience illustrates the significance of monitoring the results processed in automated campaigns and highlights an even broader challenge in automated experimentation: while the acceleration provided by automated infrastructures is highly desirable, it also enables early-stage inaccuracies to propagate into subsequent experimental decisions.
Robust data provenance and systematic data management preserved the dataset as a resource for retrospective analysis and for reuse as prior information in subsequent data-driven campaigns.
Ensuring reliability at the point of execution requires verifying not only experimental outputs, but also the computational components that analyze data and generate experimental instructions.
These considerations motivate further development towards more comprehensive validation and review mechanisms for automated scientific workflows. 
In particular, human- and AI-assisted \cite{cierpka2026kadiassistant} verification could be integrated at critical stages of the scientific process to identify inconsistencies before they propagate to experimental execution.

\section{Conclusion}
\label{sec:conclusion}
The presented framework establishes interoperability between the Kadi4Mat and \gls{FINALES} ecosystems, integrating automated and human-operated experimentation with FAIR research data management and analysis, across distributed research infrastructures.
Both systems were jointly extended to enable efficient communication and support the entire research data lifecycle. Kadi4Mat functionalities were extended through dedicated plugins for workflow coordination and data management; FINALES through tenants that connect software and laboratory services to the orchestration system and enable structured result upload.

The framework's utility and efficiency are demonstrated through its application to the automated experimental investigation of sodium-ion coin-cell formation.
The resulting experimental campaign generated a dataset covering a range of formation conditions, with batches of four cells under identical protocols providing replicated measurements and a basis for robust analysis. 
We investigated the relationship between formation parameters, formation duration, and cell longevity, to examine whether the influence of formation protocols on long-term performance, which has been demonstrated for lithium-ion cells, also extends to sodium-ion cells. 
Formation is a time-consuming and resource-intensive step in cell production. Therefore, identifying formation conditions that reduce formation duration while maintaining good long-term performance could significantly lower production costs, particularly if these insights could be transferred to larger scales.
To this end, we applied \gls{GP} modeling to describe the underlying response across the explored design space and quantify predictive uncertainty.
Although preliminary, this analysis allowed us to distinguish well-characterized regions from areas associated with greater uncertainty.
Beyond battery applications, the system's modular architecture and open-source foundation make it highly extensible and adaptable to a broad range of scientific domains.

The present framework provides a foundation for advancing from automated experimental data generation and analysis toward increasingly data-driven and autonomous experimental processes.
In future studies, we will build on the present work and reuse the generated experimental dataset as prior knowledge for data-driven multi-objective optimization. 
The selection of subsequent experimental configurations can then be delegated to an active learning agent.
Decision-making approaches that account for competing objectives can be integrated into the existing infrastructure to support more efficient allocation of experimental resources without requiring fundamental changes to the underlying interoperable architecture.
Such extensions would enable a gradual transition from automated data generation and analysis toward increasingly intelligent experimental workflows, while retaining the possibility of incorporating human expertise and oversight at selected stages of the process.

A balance between automated learning and human guidance can be maintained and tailored by systematically allocating control over distinct stages of the process to different agents, autonomous or human.
Further extensions could increase automation in the documentation and management of experimental workflows. 
For instance, an AI-assistant built on \cite{zhao2025LISALithiumIonSolidState, cierpka2026kadiassistant} could support researchers by logging errors or warnings into the Kadi record, reducing manual documentation while preserving expert oversight.


\section{Method}
\subsection{Kadi-based planning and request generation}
\label{subsec:methods-kadi}
The proposed framework operates within the Kadi ecosystem, which is built around Kadi instances that host the core repository and web frontend \cite{brandt2021kadi4mat}.
Through its modular components and plugin architecture, the Kadi ecosystem enables users to create data records and initiate events and automated workflows \cite{griem2022kadistudio}.
The constituent modules of the Kadi ecosystem communicate through well-defined interfaces, enabling seamless coordination of events and information flow. 
This design explicitly supports both automated laboratory operations and tasks performed manually by scientists and laboratory technicians.
For example, advanced visualization and dashboard-generation features make it possible to monitor and inspect accumulated data in real time (see \cref{fig_data-dashboard}).
Further supporting researchers’ daily tasks within the Kadi ecosystem is KadiStudio \cite{griem2022kadistudio}, a standalone desktop application that enables visual design and execution of scientific workflows through an interactive, graphical editor.
Within the web interface, scientific studies are represented through a network of interlinked records. 
In Kadi4Mat, a record constitutes the fundamental unit for structured data storage, combining metadata with associated files.
By default, records require only a minimal set of generic metadata (e.g., title and identifier), which is automatically enriched with system-generated information such as timestamps of the most recent changes.
A campaign can be configured using predefined templates that specify the metadata required for a particular application or research domain, including keys (labels, names), identifiers, validation rules, and predefined values.

For the present application, the user initiating a study fills in a predefined template, which automatically generates an \emph{umbrella} record.
The metadata of this record fully define the planned investigation, including design parameters, objectives (or metrics), output constraints, and settings governing data generation.
Kadi4Mat represents each described experimental batch as a trial record. 
The terms \emph{batch} and \emph{trial record} refer to the experimental and data management representations, respectively, of the same experimental iteration.
The umbrella record serves as the study's central hub: it tracks the study state throughout the process and is automatically linked to all subsequently created trial records, as shown in \ref{fig_data-knowledge-graph-kadi}.

Kadi instances provide an event-driven plugin system in which changes to records or their metadata trigger predefined actions.
Its modular architecture allows customizable plugins to implement study-specific functionalities, including data generation and interfaces to external systems. 
The framework presented here integrates the \gls{FINALES} plugin, which establishes the interface and enables communication between Kadi and \gls{FINALES}.
Both plugins automatically manage access permissions for the involved records, ensuring that only authorized users can trigger events.
These plugins automate the transfer of experimental configurations and results while maintaining the corresponding links between study, workflow, and data records.
This event-driven architecture reduces manual intervention in routine data-transfer and workflow-coordination tasks and facilitates the rapid progression from one experimental step to the next.

The workflow is initiated when the researcher marks the umbrella record with the \texttt{!umbrella\-active} tag, indicating that the study configuration is complete.
This triggers the creation of a new trial record representing the experimental configuration for the following evaluation, generated according to the data-generation method specified in the study configuration. 
Kadi can interface with different computational tools implementing diverse configuration-generation strategies, from predefined and pseudo-random to data-driven.
This allows the generation strategy to be adapted to the requirements of a given study, giving users the flexibility to choose the method that best suits their needs without imposing a specific approach.
Once generated, the configuration is stored as metadata in the trial record and automatically linked to the umbrella record via a link labeled \texttt{trial\-for}.
Two state tags are subsequently added to the trial record: a \texttt{!to\-finales}, triggering a communication request to \gls{FINALES} through the \gls{FINALES} plugin, and a \texttt{!trial\-running} once the request has been accepted.
At this point, process control is transferred to the \gls{FINALES} side, while the Kadi ecosystem waits for \gls{FINALES} to return the results.
After an experimental run is completed, the corresponding trial record is assigned the \texttt{!trial\-completed} tag, indicating successful completion.
In the event of a malfunction that prevents successful completion, a \texttt{!trial-abandoned} tag is assigned instead. 
Either of these events triggers the creation of a new trial record associated with the following evaluation, restarting the loop. 

The \gls{FINALES} plugin (developed within Kadi) provides the interface for direct communication between Kadi and \gls{FINALES}, extending the system’s interoperability with external platforms.
To define the experimental workflow, we provide a blueprint stored in Kadi as a \gls{FINALES} base workflow record and link it to the umbrella record via \texttt{!finales\-baseworkflow\-for}.
The base workflow record contains metadata extracted from a JSON file, a widely used format in web-based applications, making the \gls{FINALES} plugin easily adaptable to other services.
When the \gls{FINALES} plugin detects the \texttt{!to\-finales} tag added to a trial record, it extracts the trial configuration as key-value pairs.
The \gls{FINALES} plugin then traces the knowledge graph from the trial record to the umbrella record, and on to the base workflow record, to retrieve the corresponding workflow blueprint.
The plugin inserts the extracted key–value pairs into the blueprint and creates a \gls{FINALES} workflow record to store the experimental workflow instructions as metadata.
Finally, the plugin submits the workflow to the \gls{FINALES} server via its HTTP-based REST API, and the workflow record receives the state tag \texttt{!finales\-request\-running}.

\subsection{FINALES-based experimental workflow}
\label{subsec:methods-finales}
After successful submission via the HTTP-based REST API, the \gls{FINALES} system receives and processes the workflow, orchestrating the experiment's execution.
Within \gls{FINALES}, communication between the orchestration system and individual services, whether software or laboratory equipment, is enabled through \emph{tenants}: software clients that provide their capabilities to \gls{FINALES}.
To perform a closed-loop investigation, we adapted the previously reported \gls{FINALES} system \cite{vogler2023brokering,vogler2024autonomous} and implemented a tenant that automatically uploads the generated data to the Kadi4Mat platform upon completion.
In this study, the following tenants were registered to interact with \gls{FINALES}:
\begin{itemize}
    \item Kadi tenant -- interfaces with the Kadi4Mat platform, handling the structured upload of experimental results and metadata;
    \item OVERLORT -- the workflow manager used in the self-driving laboratory to ensure the correct sequence of events;
    \item Cycler tenant -- interfaces with the Arbin automated battery cycler;
    \item Electrolyte tenant -- an interface for a human researcher to report formulation data to \gls{FINALES}, which were collected during the manual electrolyte preparation step
    \item Transportation tenant -- coordinates the transport of physical components in the laboratory;
    \item \gls{AutoBASS} tenant -- interfaces with the Autonomous Battery Assembly System (\gls{AutoBASS}) for coin cell assembly.
\end{itemize}

\begin{figure*}[bthp]
    \centering
    \includegraphics[width=\linewidth]{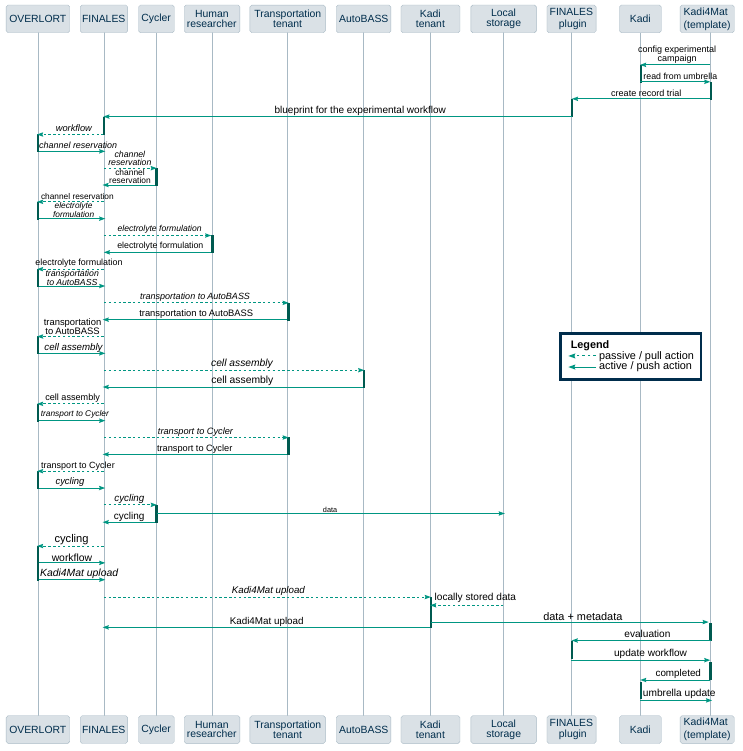}
    \caption{A representation of the overall operation of the \gls{MAP}, including also the data management tools. Active tenants constantly poll \gls{FINALES} for active requests, process them, and post the results back to \gls{FINALES}.
    Requests are set in italic and results are set in upright font.}
    \label{fig_labPart}
\end{figure*}

Each laboratory iteration starts with a workflow execution request posted to \gls{FINALES} via the \gls{FINALES} plugin in Kadi. 
The request directs the assembly and cycling of coin cells according to the specified C-rate for formation charge and discharge and the number of formation cycle repetitions.
Upon receiving the request, the \gls{OVERLORT} processes the request and orchestrates the individual tenants for both automated and manually performed tasks.
Higher automation is possible, but the executing scientists provide quality control and monitoring. 
\cref{fig_labPart} shows the sequence of events performed at the CELEST Green Energy Lab Ulm.

\begin{figure*}[bthp]
    \centering
    \includegraphics[width=0.8\linewidth]{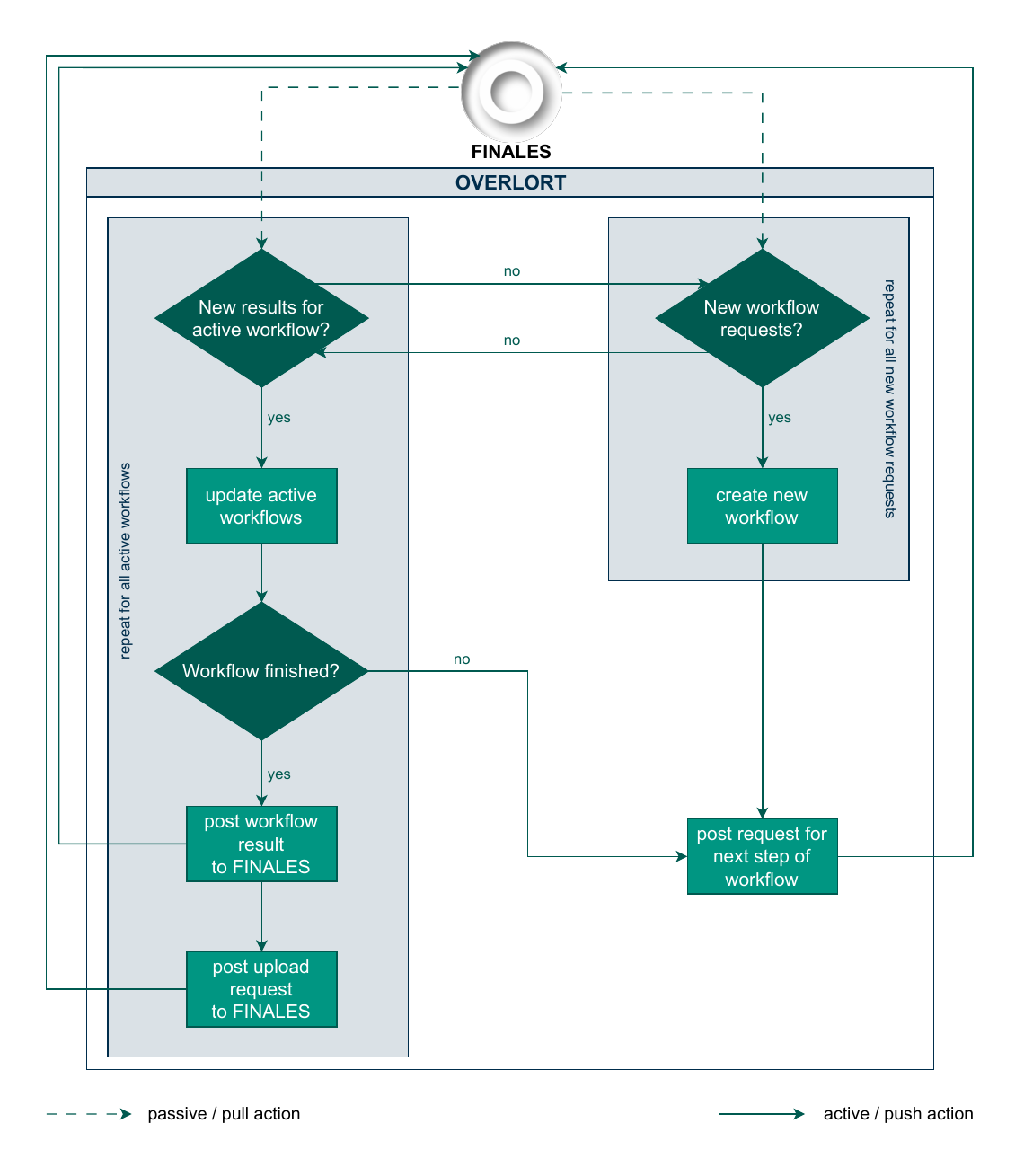}
    \caption{A representation of the process orchestrated by the \gls{OVERLORT} workflow manager. Requests for new workflows are picked up from \gls{FINALES} and split into individual requests for each process step, which are then posted back to \gls{FINALES}. The respective tenant pulls and processes these sub-requests. Once all sub-request results are available, the \gls{OVERLORT} collects them into an overall workflow result and posts it to \gls{FINALES}, along with a request to upload the data to Kadi4Mat.}
    \label{fig_OverlortWF}
\end{figure*}

After pulling the workflow request from \gls{FINALES}, the \gls{OVERLORT} posts the first sub-request and waits for the corresponding result before posting the next one.
To do so, the \gls{OVERLORT} tracks active workflows and their most recent requests in an internal queue, which is regularly updated as shown in \cref{fig_OverlortWF}.
The workflow proceeds as follows: the \gls{OVERLORT} requests the reservation of cycling channels, and the required number of channels is flagged as reserved at the battery cycler.
Subsequently, the \gls{OVERLORT} requests the electrolyte formulation. 
A researcher manually formulates the electrolyte and enters the corresponding batch data into a JSON structure that is posted to \gls{FINALES}.
In the next step, the electrolyte is manually transferred to the \gls{AutoBASS} and loaded into a vial.
The automatic cell assembly uses the provided electrolyte and further cell components, including the disk-shaped electrodes and separator.
Once the assembly is completed, the cells are transported to their reserved channels at the battery cycler outside the glovebox.
After completing the manual transfer, a researcher confirms the transport via the Transportation tenant, which closes the transport request.
The \gls{OVERLORT} submits a cycling request, which prompts the Cycler tenant to create the test protocol and start the tests at the respective channels.
Once the test is completed, the data are automatically exported, after which the \gls{OVERLORT} posts the final workflow results and a request to upload the data to Kadi4Mat. 
The Kadi tenant handles the upload request, and the workflow is removed from the internal queue.
Once the data have been evaluated, Kadi4Mat sends a new request to initiate the next workflow.

To limit the total number of experiments while accounting for experimental variability and potential outliers, we ran three batches in parallel, each comprising four cells assembled with identical materials and configuration parameters.

\subsection{Coin cell assembly and testing}
\label{subsec:methods-experimental}
Coin cells (CR2032) were automatically assembled by the in-house-developed robotic \gls{AutoBASS}-system \cite{zhang2022robotic, zhang2024apples} in a nitro\-gen-filled glovebox.
The coin cell parts were washed with isopropyl alcohol in an ultrasonic bath and dried at \SI{100}{\celsius} together with the glass fiber separator (Whatman GF/C\texttrademark). The components were subsequently stored in the glovebox until cell assembly.
Carbon-coated sodium vanadium phosphate \ch{Na3V2(PO4)3/C} (NVP/C) was used as the cathode material, and the commercial hard carbon material KURANODE\texttrademark{} Type II from KURARAY CO., LTD was used as active material in the anode.
The Institute for Applied Materials (IAM) at the Karlsruhe Institute of Technology (KIT) supplied the electrode materials and synthesized the NVP/C.
Details of the electrode preparation and characterization are provided in \cite{stuble2024powder}.
Cathodes, anodes, and separators were cut into discs of \SI{14}{\milli\meter}, \SI{15}{\milli\meter}, and \SI{16}{\milli\meter} diameter, respectively.
\SI{70}{\micro\liter} of \SI{1}{\molr}~\ch{NaPF6} in ethylene carbonate (EC): propylene carbonate (PC) (1:1~by weight) was used as the electrolyte.
The electrolyte was manually formulated in batches large enough to be used for several batches of cells.
During the cell assembly process, the \gls{AutoBASS}~\cite{zhang2022robotic, zhang2024apples} dosed the electrolyte into the coin cells using its pipetting module.

The nominal capacity for all cells was calculated to be \SI{1.48}{\milli\ampere\hour} based on the areal cathode capacity of \SI{0.97}{\milli\ampere\hour}\linebreak\si{\per\square\centi\meter}.
All C-rates reported in this study are referenced to this nominal capacity, which was used to determine the charging and discharging currents for each batch.
The cells were cycled at a controlled temperature of \SI{20}{\celsius} on an Arbin battery cycler (model LBT21084-5).
The test protocols include formation cycle(s) at various C-rates and repetitions, as well as cycling using constant current constant voltage (CCCV) charging at C/5 up to \SI{3.9}{\volt} (C/20 cut-off) and C/5 discharging until the cell voltage falls below \SI{2.3}{\volt}. 
The cycling was repeated until a threshold of \SI{80}{\percent} of the remaining discharge capacity was reached, using the first cycle as the reference.

\subsection{Closing the loop with the human}
\label{subsec:methods-closing-loop}
Back on the Kadi side, the \gls{FINALES} plugin identifies the result records uploaded by the \gls{FINALES} Kadi tenant.
Two approaches are possible. First, a metadata field in the workflow record stores the request \gls{UUID} returned by the \gls{FINALES} instance once a request is posed. This \gls{UUID} can be used to search for uploaded result records containing the same \gls{UUID}, including records stored across different Kadi instances. 
Alternatively, the \gls{FINALES} plugin detects result records through links added to the workflow record containing the request.
In the present study, we used the first option because development was performed on a dedicated test Kadi instance, whereas \gls{FINALES} communicated results from a production Kadi instance.

Based on the result records, the \gls{FINALES} plugin extracts and evaluates cycling data and associated parameters.
A Python script analyzes the cycling curves for each cell within each batch, extracts characteristic points and computes the mean and \gls{SEM} for the set objectives.
Because actual experimental conditions may deviate from those specified in the original workflow request due to intrinsic device variability, experimental errors, or human intervention, the data points are updated with the measured design parameters.
Finally, the \gls{FINALES} plugin saves the updated design parameters and metrics of interest in the workflow and trial records, respectively, and updates their state tags to \texttt{!finales\-request\-completed} and \texttt{!trial\-completed}.

The loop can be closed either manually or by the plugin system, which detects completion of new trials.
Upon this event, the plugin retrieves and aggregates the data from all completed trials associated with the corresponding study and the related umbrella record.
The measured parameter configurations, together with the corresponding performance metrics and variances, are stored and subsequently used to fit \gls{GP} models.

\begin{figure*}[bthp]
    \centering
    \includegraphics[width=\linewidth]{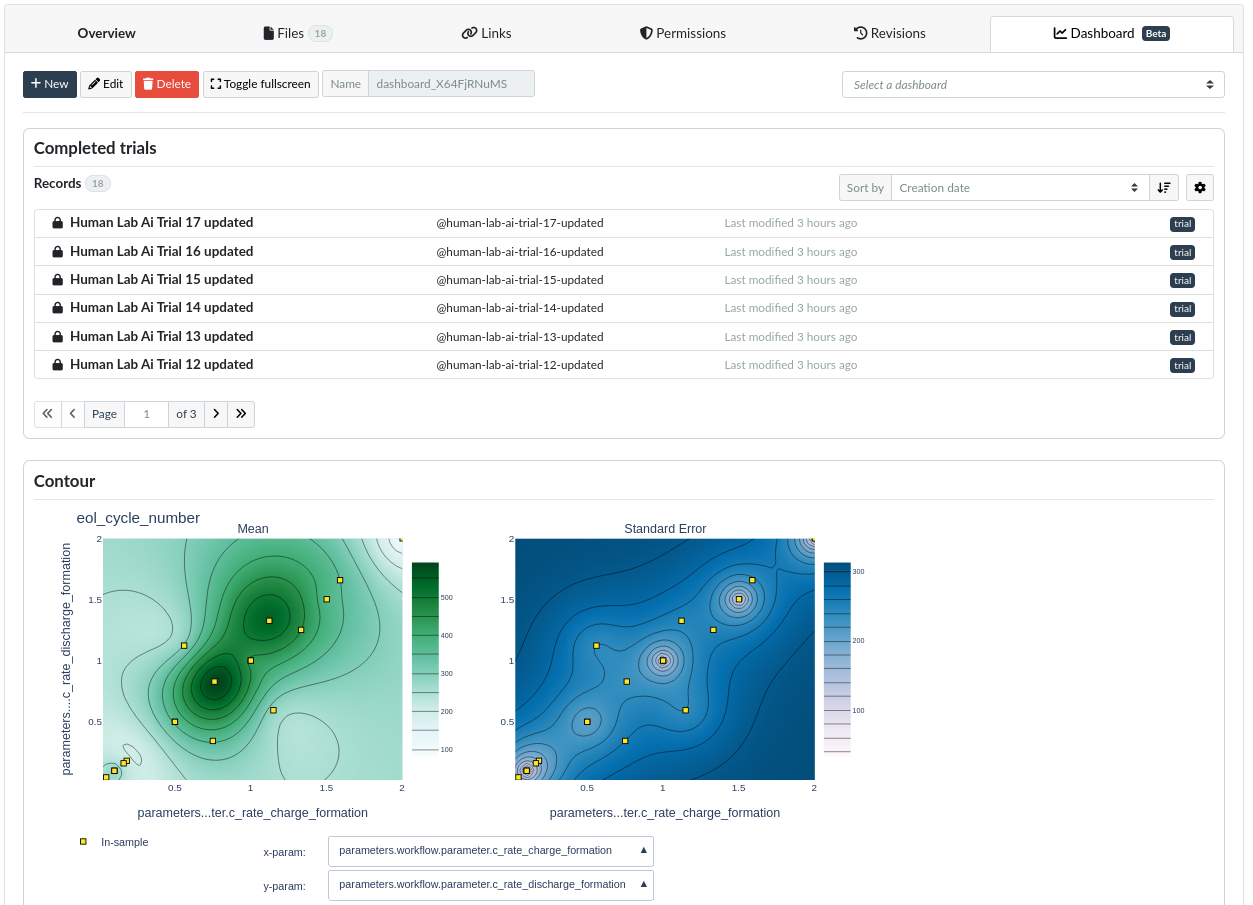}
    \caption{The dashboard visualization provides the scientists with an overview of the study. The panels and plots can be rearranged, adapted, replaced, and extended. The dashboard shown here displays the most recently completed trials together with automatically generated contour plots representing the model's estimate across the performance space and its associated uncertainty.}
    \label{fig_data-dashboard}
\end{figure*}

To provide an overview of the study and its accumulated experimental data, umbrella records include optional dashboards (\cref{fig_data-dashboard}).
The dashboards provide interactive visualization of the data and their distribution across the investigated parameter domain, including contour plots of the modeled responses, updated after each iteration. 
This makes the evolving dataset, along with the corresponding model predictions and uncertainties, accessible throughout the study.
Internally, the dashboards are implemented in Python through the Dash and Plotly libraries, which export interactive plots as JSON objects that can be embedded directly in the Kadi web interface.

\subsection{Data structure and knowledge graph}
\label{subsec:methods-datastructures}
In \gls{FINALES}, all services, whether physical setups in a laboratory or computations on a server or computing cluster, are registered as tenants.
Upon registration, \gls{FINALES} assigns to each tenant a \gls{UUID}.
Before registration, the tenant developer and the \gls{FINALES} administrator define the data structure for input and output exchanged with the tenant.
All requests and results posted to \gls{FINALES} are validated against these data structures and rejected if they are not compliant.
Each tenant provides to the \gls{MAP} one or more capabilities, defined as a combination of a certain quantity and a method that generates a value for it.
In the scope of \gls{FINALES}, a quantity is not limited to a measurable, physical property, but can also be defined as a label for the output of a service (e.g., transportation).
Tenants accesses the \gls{MAP} by querying \gls{FINALES} and submitting either requests or results, depending on the assigned task.

All requests and results posted to \gls{FINALES} are stored in an SQL database.
Upon submission, \gls{FINALES} assigns a \gls{UUID} to each request and result to enable traceability and link results to their corresponding requests.
For each request, \gls{FINALES} also records a timestamp to document the receiving time, and assigns it the status \texttt{pending}.
The specific status label enables tenants to identify and retrieve tasks by regularly querying \gls{FINALES} for pending requests that match their capabilities.
In addition, each request object contains all the parameters actually used during the execution of methods and tasks that generated the requested data. 

A tenant that picks up a request performs the corresponding tasks using the specified inputs and generates a result. Alternatively, it may process results previously made available through \gls{FINALES} and post a new request.
Results posted to \gls{FINALES} share several structural characteristics with requests. 
Both contain the input parameters established for task execution, and their bodies contain the data obtained by the tenant in the specified output format.
In addition, both are automatically assigned metadata, such as \gls{UUID},  timestamp, and status, upon posting.
A key distinction is that each result also includes the \gls{UUID} of the request it replied to, thereby establishing a link between the request and its result.
Furthermore, the result records the parameters the tenant actually used during task execution.
This allows for the comparison of targeted and actual parameter values, which may, for example, be used to identify severe discrepancies. 

Each registered tenant has its own data structures for inputs and outputs, based on its specific capabilities.
These structures are specified using JSON schemas, which both humans and programs interfacing with \gls{FINALES} can read and process.
\gls{FINALES} provides an endpoint for querying the data structures registered for a specific capability.
In addition, it was decided not to store data files in a binary format within the SQL database, although this is technically possible.
Within the SQL database of \gls{FINALES}, links between requests, results, tenants, and other objects are traceable using the \gls{UUID}s assigned to each element in the database.
Although this structure provides comprehensive machine-readable traceability, the absence of a graphical representation might make it challenging for a human researcher to quickly grasp these relations and get an overview of the interconnected workflow.

Kadi4Mat complements this functionality by providing researchers with a more accessible and intuitive overview of the workflow and its associated data.
The Kadi4Mat tenant not only uploads data for persistent storage but also organizes them into a structured hierarchy of collections, records, and links.
These relationships can be visualized within Kadi4Mat, facilitating navigation through the data and providing a more intuitive representation of the connections between experimental activities and their associated information.
In this study, the structure created in Kadi4Mat organized around two hierarchical levels: a \emph{campaign} and one or more associated \emph{studies}.

A campaign provides a high-level organizational structure (e.g, a research project or a research group) and can encompass multiple related studies.
Kadi4Mat represents campaigns and studies as collections and associated records, respectively.
A new campaign begins when the administrator creates the corresponding collection and record, which serve as the organizational basis for the studies conducted within it.
Studies can be linked to a campaign based on their scientific or organizational relationship, for example, through shared funding.
The campaign considered in this study is named \emph{Auto-POLiS}, and contains the \emph{Human-Lab-AI} study. 
The corresponding record and collection in Kadi4Mat are shown at the center of \cref{fig_data-knowledge-graph-finales}. 

A reduced view of the knowledge graph, focusing on the records associated with the generated experimental configurations, is shown in \cref{fig_data-knowledge-graph-kadi}.
The umbrella record at the center holds the complete study configuration and connects to the \gls{FINALES} interface.
It is linked to the individual trial records, each of which contains the data and metadata for a specific batch of cells.
Each trial record is linked to a \gls{FINALES} request record, which communicates with \gls{FINALES} and triggers the corresponding workflow. 
The user can control the workflow through tags on the request record, allowing it to be started, stopped, or evaluated.
Records can also be linked to one another, explicitly representing relationships between different entities within a study.

As shown in \cref{fig_data-knowledge-graph-finales}, the workflow records form an outer layer around the study record.
These records represent the workflows executed by the \gls{OVERLORT} in response to requests submitted through the \gls{FINALES} plugin.
The results generated during workflow execution are stored in individual records in Kadi4Mat and linked to their corresponding workflow record.
These include results for reserving cycling channels via the \emph{service} method, electrolyte formulation, and executed transports, assembly, and cycling.
Since coin cell assembly is requested in batches of four cells, each workflow includes one record labeled \texttt{autobass\_assembly} and four distinct records corresponding to cycling the individual cells.

After each workflow request is posted to \gls{FINALES}, the data is uploaded to Kadi4Mat, where it is organized into a linked set of records.
The graphical representation of the structures shown in \cref{fig_data-knowledge-graph-finales} provides researchers with an overview of the study and the connections between the results.
The Kadi4Mat tenant also assigns a designated user group as administrators of each record, facilitating data sharing and access management by simply adding or removing users from the respective group.
When different studies are associated with distinct user groups, this mechanism enables straightforward permission management and fine-grained control over data access.

Since the data are provided in structured, machine-readable formats in \gls{FINALES} and Kadi4Mat, mappings to ontologies such as BattINFO \cite{Clark2022Toward, Clark2023BattINFO} can be created.
Future versions of \gls{FINALES} could implement mappings to ontologies either directly in the \gls{FINALES} schemas or through tenants that map predefined schema keys to ontology terms.
The latter option allows multiple tenants to deploy mappings to different ontologies.
Such functionality could also be included in tenants, such as the Kadi4Mat tenant, if a platform requires a specific ontology.


\section{Usage of artificial intelligence}
Generative AI (Grammarly) was used to improve the manuscript's readability, grammar, and spelling.

\section{Data availability statement}
This research used open-source Python software libraries, including Ax \cite{olson2025ax} and CIDS \cite{koeppe2023cids}.
The data generated during this study is publicly available in the Zenodo repository (https://zenodo.org/records/19360411). 
The Zenodo record also includes a link to the specific version of the source code available in the GitLab repository.

\section{Acknowledgements}

This work contributes to research at the CELEST (Center for Electrochemical Energy Storage) Green Energy Lab Ulm and was funded by the German Research Foundation (DFG) under Project ID 390874152 (POLiS Cluster of Excellence) and by the Helmholtz Association within the program MSE no. 43.31.01.

The authors acknowledge the collaborators from IAM-ESS at the Karlsruhe Institute of Technology (KIT) for providing the  \gls{POLiS} reference electrodes and information about these materials. The authors further acknowledge Andreas Hofmann for providing the \ch{NaPF6} salt and for his advice in formulating the electrolyte.
The Thin Film Technology (TFT) group at KIT (especially Julian Klemens) is acknowledged for preparing the \gls{POLiS} reference anodes.
The authors also acknowledge the support of Usman Hayder, Prashant Sai Bharadwaja N V and Adam Reupert in assembling and cycling the coin cells.

The authors would like to thank the German Federal Ministry of Research, Technology and Space and the German federal states (https://www.nhr-verein.de/nhr-alliance/ governance/) for supporting this research project as part of the National High-Performance Computing (NHR) joint funding program.

\section{Conflict of Interest}

The authors declare no competing interests.

\begin{shaded}
\noindent\textsf{\textbf{Keywords:} \keywords} 
\end{shaded}


\setlength{\bibsep}{0.0cm}
\bibliographystyle{Wiley-chemistry}
\bibliography{references}


\newpage
\setcounter{figure}{0}    
\renewcommand{\thefigure}{S\arabic{figure}}  

\section*{Supporting Information}

\begin{figure*}[bthp]
    \centering
    \begin{subfigure}[b]{0.8\linewidth}
        \centering
        \includegraphics[width=0.7\textwidth]{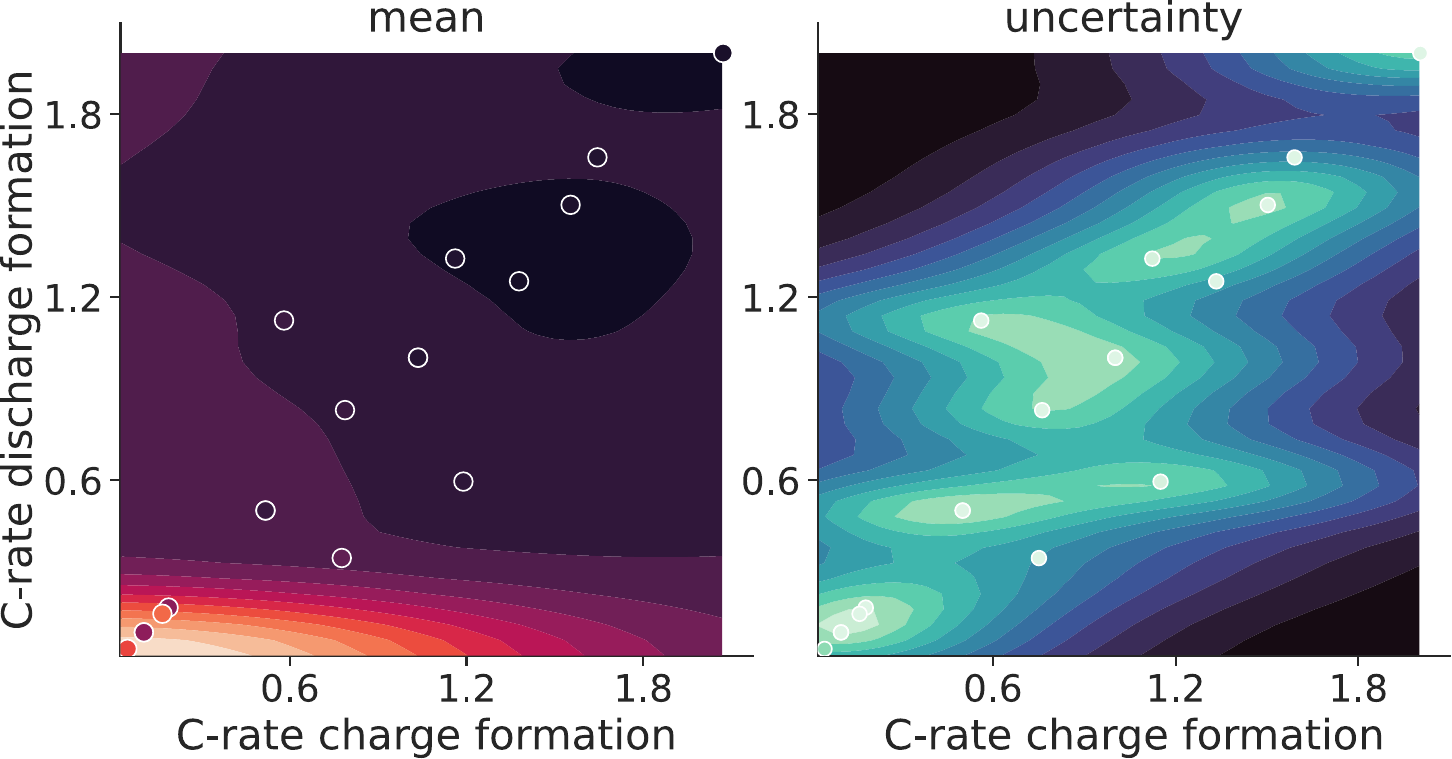}
        \caption{}
    \end{subfigure}
    \par\bigskip 
    \begin{subfigure}[b]{0.8\linewidth}
        \centering
        \includegraphics[width=0.7\textwidth]{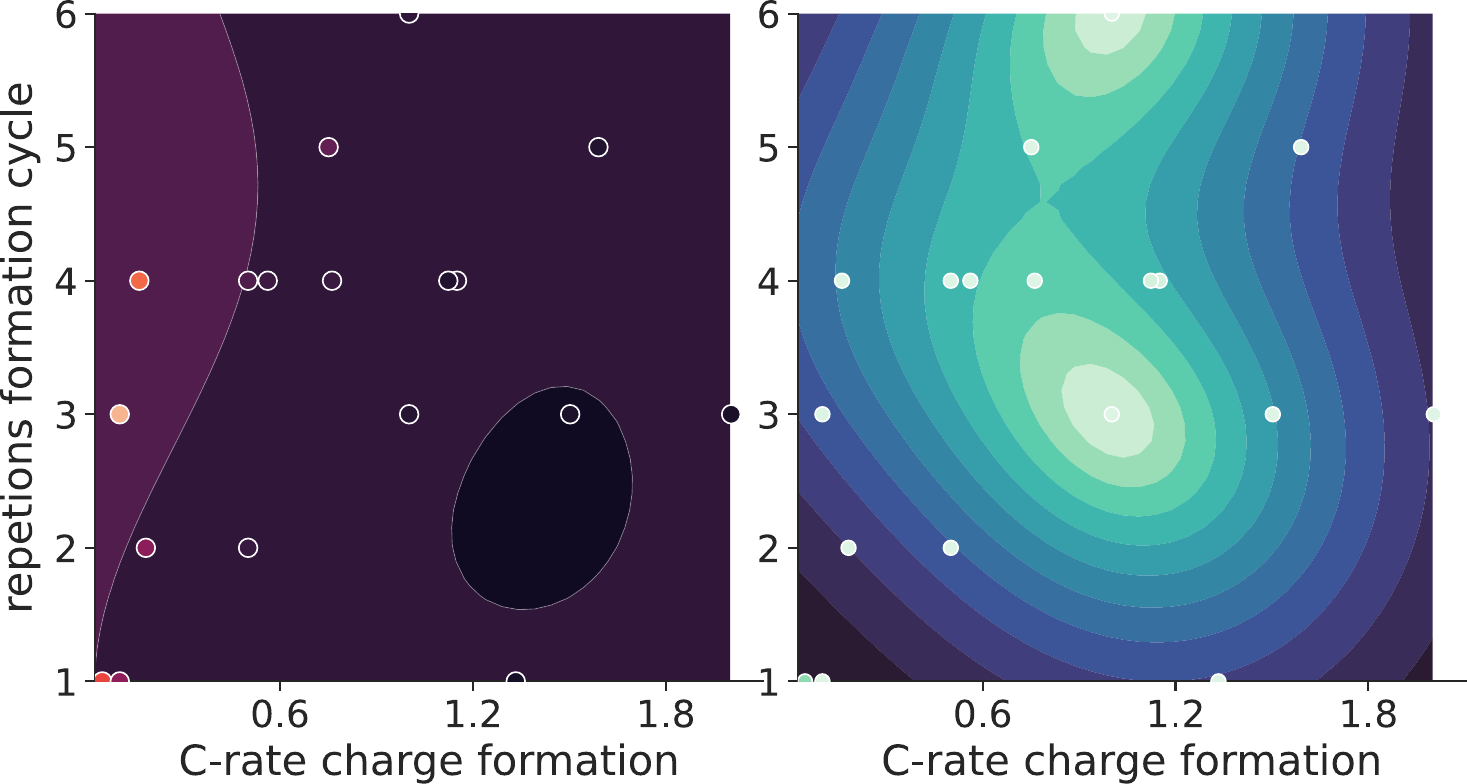}
        \caption{}
    \end{subfigure}
    \par\bigskip 
    \begin{subfigure}[b]{0.8\linewidth}
        \centering
        \includegraphics[width=0.7\textwidth]{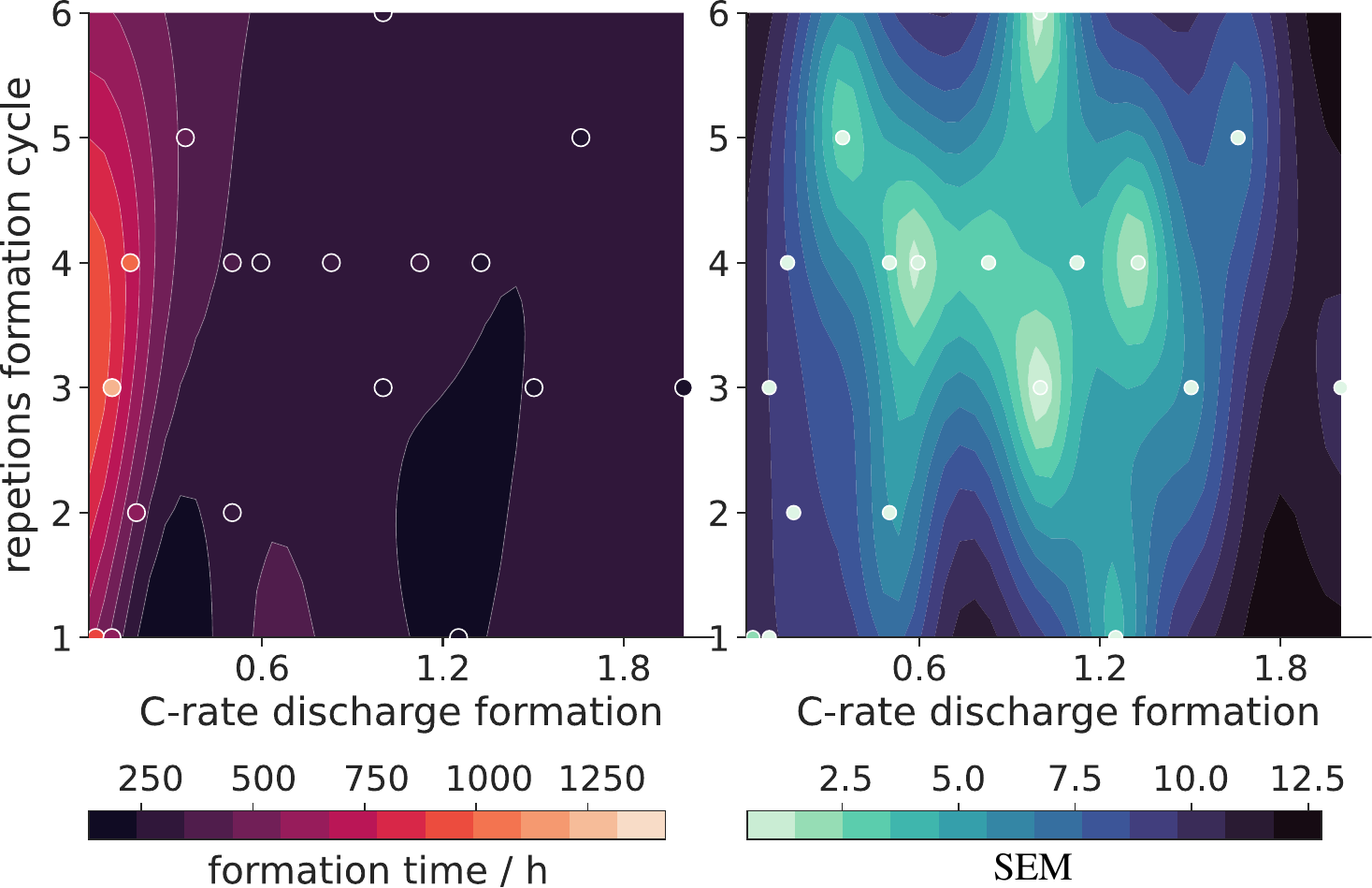}
        \caption{}
    \end{subfigure}
    \caption{Posterior mean and standard error of the \gls{RBF} \gls{GP} for formation time across the investigated cycling-protocol parameter space. Contour plots show the modeled formation time response for the three pairs of input dimensions: (a) charge and discharge rate, (b) charge rate and number of repetitions, (c) discharge rate and number of repetitions.}
    \label{fig_formation_contour}
\end{figure*}

\begin{figure*}[bthp]
    \centering
    \begin{subfigure}[b]{\linewidth}
        \centering
        \includegraphics[width=\textwidth]{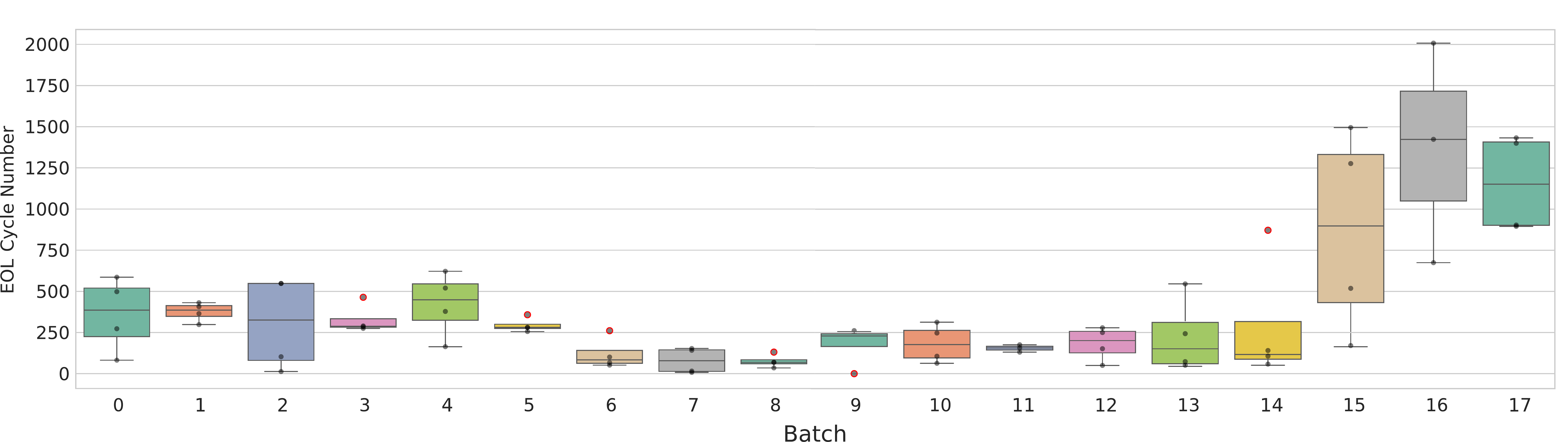}
        \caption{}
        \label{fig_eol_boxplot}
    \end{subfigure}
    \begin{subfigure}[b]{\linewidth}
        \centering
        \includegraphics[width=\textwidth]{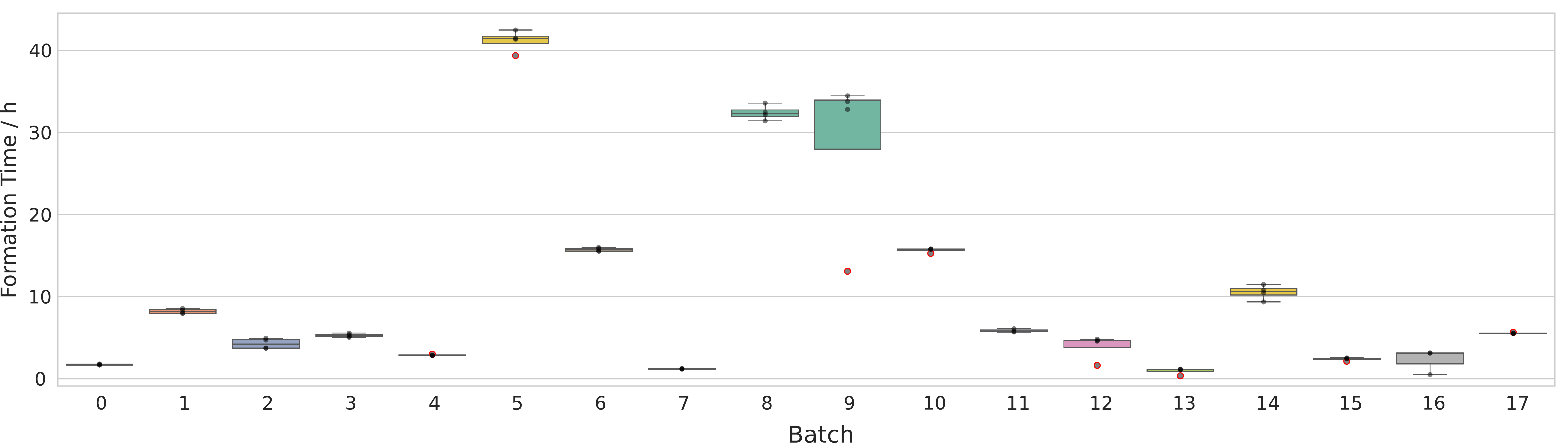}
        \caption{}
        \label{fig_formation_time_boxplot}
    \end{subfigure}
    \caption{Box plots of (a) \gls{EOL} and (b) formation time cycle number for all 18 batches. The distribution of the four cells per batch: the horizontal line indicates the median, the box spans the interquartile range, and whiskers extend to the minimum and maximum non-outlier values. Individual cell values are overlaid as black points; outliers are marked with open red circles. The high-\gls{EOL} batches (15, 16, 17) (a) exhibit substantially large cell-to-cell dispersion, reflecting the increased experimental uncertainty in this performance region. Formation time (b) shows low cell-to-cell variability across all batches, consistent with it being a near-deterministic function of the protocol parameters.}
    \label{fig_boxplots}
    \end{figure*}

\begin{figure*}[bthp]
    \centering
    \includegraphics[width=\linewidth]{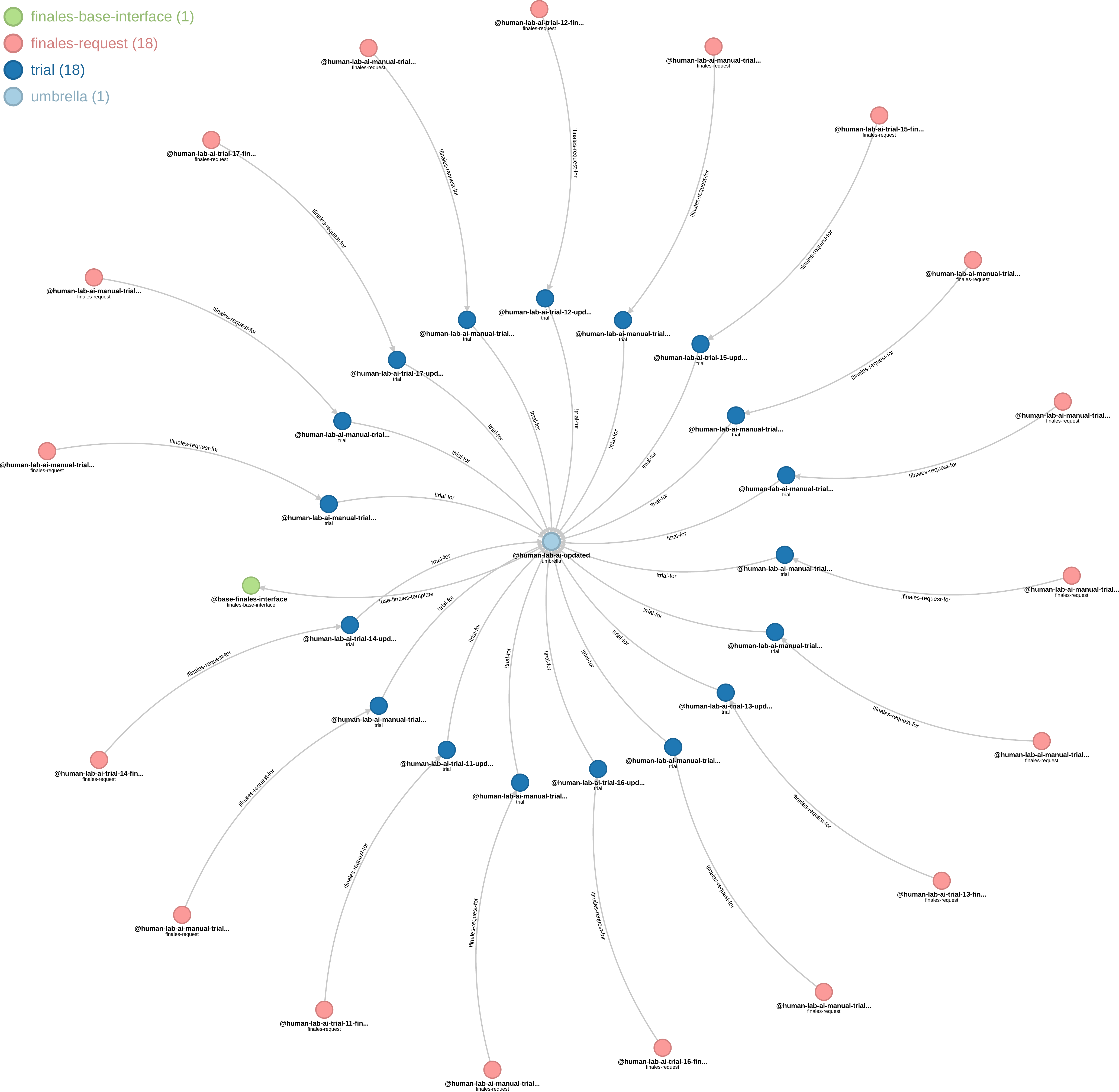}
    \caption{Reduced view of the knowledge graph showing the records of the experimental configurations generated during the study. The umbrella record (in light blue at the center), is connected to all trial records (dark blue), each one representing one experimental batch and containing its corresponding parameter configuration. Each trial record is linked to a \gls{FINALES} request record (pink), used to trigger the corresponding \gls{FINALES} workflow.}
    \label{fig_data-knowledge-graph-kadi}
\end{figure*}

\begin{figure*}[bthp]
    \centering
    \includegraphics[width=\linewidth]{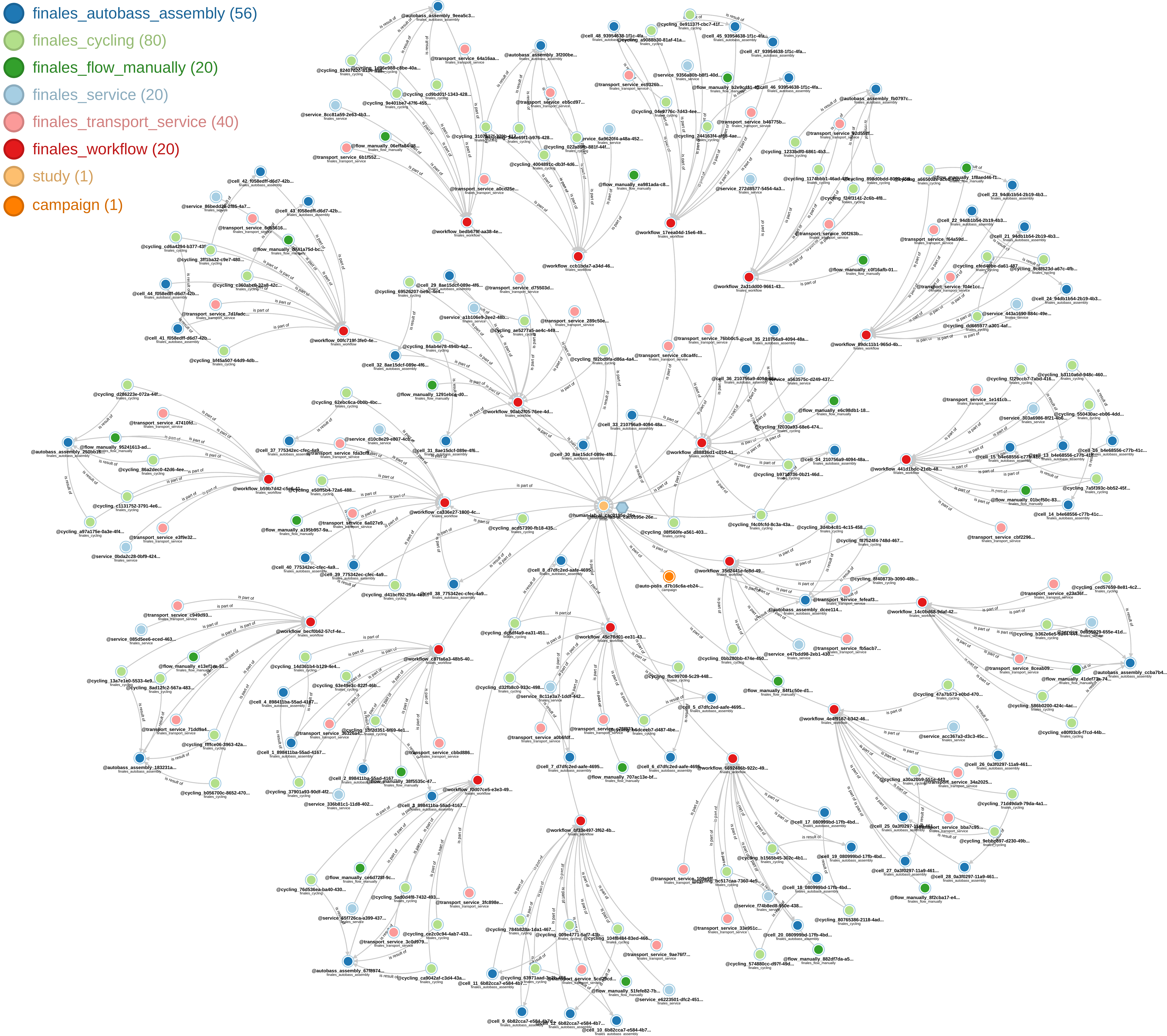}
    \caption{The full knowledge graph from the \gls{FINALES} perspective.}
    \label{fig_data-knowledge-graph-finales}
\end{figure*}

The Supporting Information provides supplementary visualizations that complement the main text. 
\cref{fig_formation_contour} presents the \gls{GP} posterior for formation time, \cref{fig_boxplots} displays cell-to-cell variability within each batch for both descriptors, and \cref{fig_data-knowledge-graph-kadi,fig_data-knowledge-graph-finales} show the knowledge graph structures in Kadi4Mat.

\end{document}